\documentclass[10pt,twocolumn,letterpaper]{article}
\usepackage[accsupp]{axessibility}
\usepackage{wacv}
\usepackage{times}
\usepackage{epsfig}
\usepackage{graphicx}
\usepackage{amsmath}
\usepackage{amssymb}
\usepackage{booktabs}

\usepackage{xcolor}
\usepackage{array}
\usepackage{subfigure}
\usepackage{epsfig}
\usepackage{epstopdf}
\usepackage{booktabs}
\usepackage{algorithm}
\usepackage{algpseudocode}
\usepackage{lipsum}
\usepackage{multirow}
\usepackage{textcomp}
\usepackage{gensymb}
\usepackage{pifont}

\newcommand{\cmark}{\ding{51}}%
\newcommand{\xmark}{\ding{55}}%

\def\wacvPaperID{1188} %

\wacvalgorithmstrack   %

\wacvfinalcopy %

\newcommand{\HD}[1]{{\color{black}#1}}

\ifwacvfinal
\usepackage[breaklinks=true,bookmarks=false]{hyperref}
\else
\usepackage[pagebackref=true,breaklinks=true,colorlinks,bookmarks=false]{hyperref}
\fi

\begin{document}

\def\papertitle{Gait Recognition Using 3-D Human Body Shape Inference}

\title{\papertitle}
\author{Haidong Zhu\ \ \ \ \ \ Zhaoheng Zheng\ \ \ \ \ \ Ram Nevatia\\
University of Southern California\\
{\tt\small \{haidongz|zhaoheng.zheng|nevatia@usc.edu\}}
}
\maketitle
\thispagestyle{empty}

\begin{abstract}
Gait recognition, which identifies individuals based on their walking patterns, is an important biometric technique since it can be observed from a distance and does not require the subject's cooperation. Recognizing a person's gait is difficult because of the appearance variants in human silhouette sequences produced by varying viewing angles, carrying objects, and clothing. Recent research has produced a number of ways for coping with these variants. In this paper, we present the usage of inferring 3-D body shapes distilled from limited images, which are, in principle, invariant to the specified variants. Inference of 3-D shape is a difficult task, especially when only silhouettes are provided in a dataset. We provide a method for learning 3-D body inference from silhouettes by transferring knowledge from 3-D shape prior from RGB photos. We use our method on multiple existing state-of-the-art gait baselines and obtain consistent improvements for gait identification on two public datasets, CASIA-B and OUMVLP, on several variants and settings, including a new setting of novel views not seen during training.

\end{abstract}

\section{Introduction}

\begin{figure}[t]
    \centering
    \includegraphics[width=\linewidth]{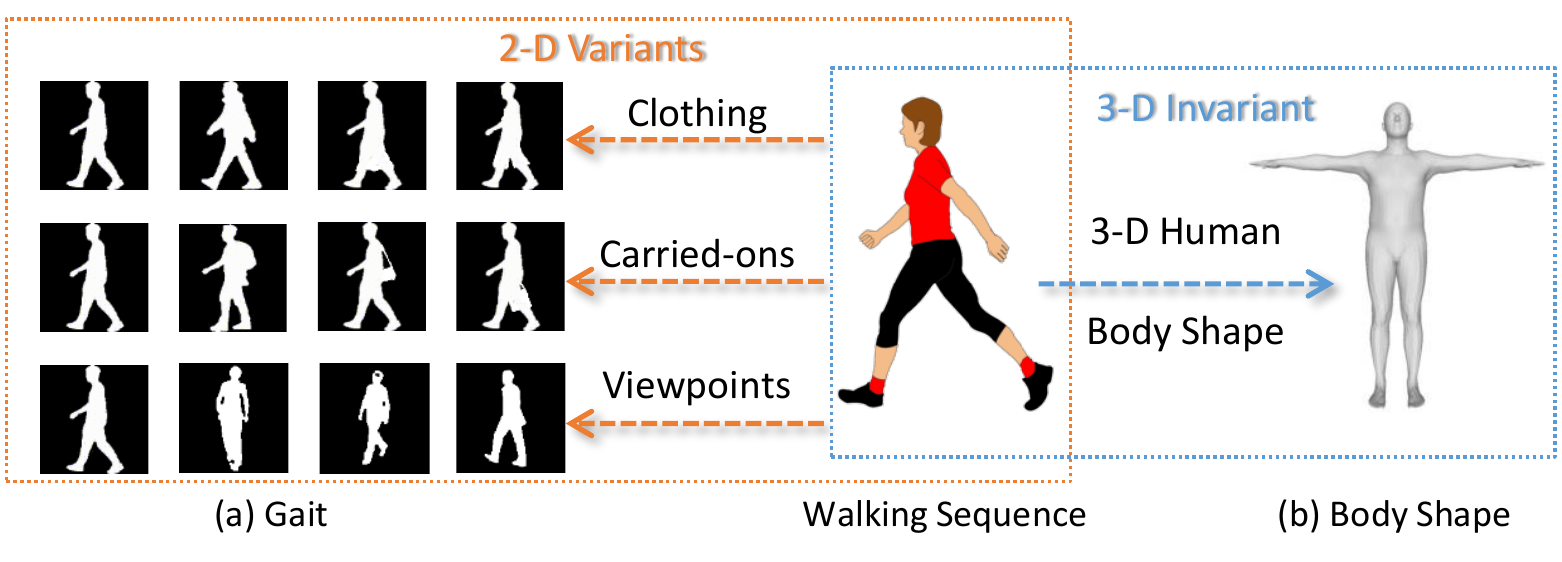}
    \caption{(a) 2-D silhouette sequences suffer from different appearance variances, such as clothing, carried-on bags and camera viewpoints. (b) However, 3-D skinned human body shape is robust and shows consistent output for the same person.}
    \label{fig:example}
\end{figure}

Many biometrics, such as face ID \cite{deng2019arcface,schroff2015facenet}, have been developed for automated human identification. 
One such biometric is gait, which has the advantage of being able to be acquired from long distance and without the subjects' cooperation. 
Gait recognition \cite{he2018multi,song2019gaitnet,wu2016comprehensive,yu2017invariant} aims to find the uniqueness for a sequence of walking patterns and posture of a person in the binarized silhouette sequence describing human boundaries. 
However, appearance variances in 2-D images, like camera positions, carried-on objects, and clothing, introduce additional disparity in the human shape and make the task of recognition challenging. 
Fig.~\ref{fig:example}~(a) shows these variations in extracted silhouette for a person under three different appearance variances. 

To address the issue of appearance variances, researchers have developed part-based deep-learning models that focus on local patterns.
For example, GaitPart~\cite{fan2020gaitpart} splits the image into several patches to encode the part-based features for gait recognition, whereas GaitGL~\cite{lin2021gaitgl} utilizes local features along with the global ones for the analysis.
By limiting the variances to a small portion of the feature, these strategies aim to reduce the influence of the variances.
However, features encoded by these approaches are still impacted. 

In this paper, we propose inferring 3-D human body shape representations directly from silhouette sequences by using knowledge distillation to learn from a small number of RGB samples.
We observe that the 3-D body shape of the human, as illustrated in Fig.~\ref{fig:example}~(b) for example, is, in principle, \HD{invariant to viewpoint, carrying objects, and clothing, and might therefore be useful for human identification under such challenging scenarios.}
Nevertheless, inference of the underlying 3-D shape is difficult in and of itself. Recent work \cite{saito2019pifu,SMPL-X:2019,zheng2022gait} has created numerous approaches to infer 3-D shape from RGB images, but no work directly infers the body shape from a silhouette sequence. Many gait datasets lack RGB photos due to confidentiality, making such inferences more difficult.

To infer 3-D body shapes from the silhouette sequence, we exploit a temporal shift between the features obtained from adjacent frames in the silhouette sequence. Considering the consistency of the body shape of the same individual in a video sequence, we extract and reconstruct a single body shape for a video sequence.
\HD{We combine body shapes acquired from our approach with 2-D gait features collected from certain state-of-the-art gait recognition methods \cite{chao2019gaitset,fan2020gaitpart,lin2021gaitgl,hou2020gln} to build our module on each of them.} To supervise the generation of the body shape from the silhouette sequence, we distill and transfer the knowledge from a small set of RGB images, \HD{denoted as human body prior}, and propagate it to gait. We demonstrate that adding 3-D body shape feature inferred from silhouette sequence significantly improves gait recognition accuracy on two public datasets (CASIA-B~\cite{yu2006framework} and OUMVLP~\cite{takemura2018multi}), particularly for novel viewpoints that were not observed during training \HD{with fewer available training instances}, which is a new setting in our experiment.

A recent paper, Gait3D \cite{zheng2022gait}, has also proposed using 3-D body shape for gait recognition. \HD{However, our work differs in the following manner:} instead of inferring 3-D body shapes from all RGB frames, we infer 3-D shapes from silhouettes via distilling knowledge from a small set of images. Another is in our use of temporal information in 3-D inference. \HD{Gait3D extracts framewise body shape, while we extract video-level body shape using temporal consistency.} %

In summary, our contributions are summarized as follows: 1) We apply the 3-D human body inferred from gait to eliminate the effects of \HD{different clothing, carried-on objects and viewpoints} for gait recognition; 2) We distill the knowledge of human body \HD{prior} from limited single-frame RGB images and transfer to the silhouette sequence for body shape reconstruction directly from gait; and 3) \HD{We explore the setting for gait recognition on novel camera positions to assess the generalization of gait recognition models with fewer available data.} %

\section{Related Work}

\indent \textbf{Gait Recognition.}
For a silhouette sequence describing a person's walking pose, gait recognition is to find the corresponding identity of the person in the gallery. 
Recently, researchers have proposed different methods \cite{teepe2021gaitgraph,song2019gaitnet,chao2019gaitset,lin2021gaitgl,hou2020gln,fan2020gaitpart,zheng2022gait,liang2022gaitedge,fan2022learning,zhu2021gait,li2020end} for extracting the identity information from the gait sequence for recognition. 
GaitSet~\cite{chao2019gaitset} treats the whole sequence as a set of different images for set pooling and feature fusion. 
GaitPart~\cite{fan2020gaitpart} splits the gait image into different parts and extracts the feature from each local pattern for analysis. GLN~\cite{hou2020gln} utilizes both silhouette-level and set-level features and fuses them for different gait analyses at different stages.
GaitGL~\cite{lin2021gaitgl} introduces using both local and global features:  \HD{local features are computed by splitting an image into several patches and encoding the feature for each patch}; global features \HD{are framewise encoded features} and combine them together for gait recognition. 
GaitNet~\cite{song2019gaitnet} and GaitGraph~\cite{teepe2021gaitgraph} use the consistency between RGB images and graph recognition network for recognizing the identity of the human in the sequence. 
These methods focus on extracting and distinguishing information directly from the 2-D gait sequence. \HD{Other methods, such as PoseGait \cite{liao2020model} and ModelGait \cite{li2020end}, require the RGB images for all the training instances, which are sometime difficult to get due to privacy issues.}

{Gait3D~\cite{zheng2022gait} uses the 3-D body shape extracted from RGB images, which has extra input compared with other methods.}
Since the gait sequences are binarized images, when people have different clothing or carried-on objects and are shot by the camera from different positions, predictions from the features extracted are affected.

\textbf{Knowledge Distillation.} 
The task of knowledge distillation is to transfer the knowledge from one model to another. 
Knowledge transfer has been successfully applied in tasks such as visual and speech recognition~\cite{he2019bag,chebotar2016distilling} and between different modalities~\cite{tian2019contrastive}, etc. 
Researchers have proposed several methods for knowledge distillation and transfer~\cite{ahn2019variational,heo2019knowledge,hinton2015distilling,huang2017like,kim2018paraphrasing,park2019relational,yim2017gift,zagoruyko2016paying}. 
For these methods, their inputs for different models are mostly from the same modalities: both the source and target are data sequences or single frames. 
Knowledge distillation and transfer from RGB images to gait sequences require understanding both gait sequences and single-frame images. 
To transfer the knowledge from an image to a video, we need to distribute the information to individual frames of the video.

\textbf{3-D Body Shape Reconstruction.} 
A model needs lots of the knowledge \cite{hu20213dbodynet,jin2022grouping} to reconstruct the 3-D body shapes. 
Methods for 3-D body shape reconstruction can be divided into two mainstreams: parametric methods, such as SMPLify-X~\cite{SMPL-X:2019} and SMPLify~\cite{bogo2016keep}, and non-parametric methods, such as PIFu~\cite{saito2019pifu} and PIFuHD~\cite{saito2020pifuhd}. 

For parametric methods, SMPLify~\cite{bogo2016keep} and SMPLify-X~\cite{SMPL-X:2019} reconstruct the human body shape based on the pre-defined parameterized skinned models, SMPL~\cite{loper2015smpl} and SMPL-X~\cite{SMPL-X:2019}. 
As non-parametric methods, PIFu~\cite{saito2019pifu} and PIFuHD~\cite{saito2020pifuhd} utilize the implicit function for representing the shape and predict whether points in the space are inside or outside the object.
These non-parametric methods do not record locations of points on the object surface but understand the whole body shape correspondingly. 
With the introduction of NeRF~\cite{mildenhall2020nerf}, researchers also introduce Animatable NeRF~\cite{peng2021neural} and Neural Body~\cite{peng2021neural} for reconstructing the human body shape in the video sequence with SMPL priors. 
Compared with these methods, due to the lack of RGB images in gait datasets, we use the distilled body prior from a small set of examples and extract body shapes from the silhouette sequence instead of RGB images.

\section{Method}

\begin{figure*}[t]
    \centering
    \includegraphics[width=0.9\linewidth]{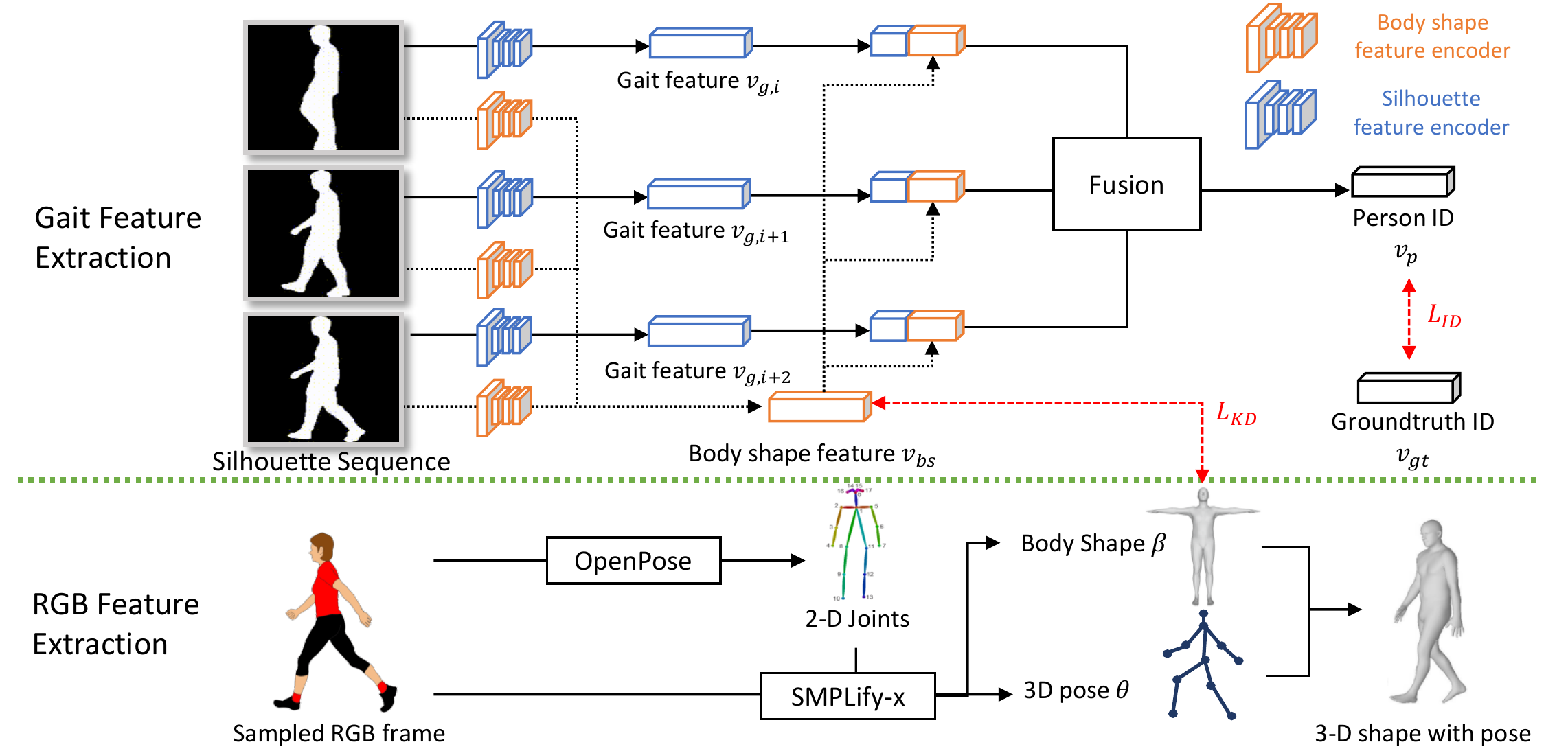}
    \caption{Our Proposed method for gait recognition with 3-D human body shape. During inference, only the features extracted from gait branch are used. Features from RGB images (below the green line) are only used for training when corresponding images are available.}
    \label{fig:pipeline}
\end{figure*}

Our network consists of two branches, one for gait feature extraction and the other for body shape feature extraction from RGB images, which is shown in Fig.~\ref{fig:pipeline}.
For gait inputs, we have a silhouette feature encoder and a body shape feature encoder to encode the gait and body shape feature.
To supervise the generation of body shapes, we simultaneously extract knowledge from selected RGB frames using a human body reconstruction model.
Then, we extract and transfer these inferred body shape information from RGB frames to the gait branch's body shape encoder. 

In the remaining of this section, we will first introduce our gait pipeline in Sec.~\ref{sec:gait} for how gait features and body shape features are extracted for identification, and then discuss how body shape of selected RGB images are generated and used as the supervision for the gait branch in Sec.~\ref{sec:transfer}.

\subsection{Gait Feature Extraction}\label{sec:gait}
We propose two feature encoders to extract features from gait images: a silhouette feature encoder to extract walking patterns from the gait sequence and a body shape feature encoder to extract the body shape.

\textbf{Silhouette Feature Encoder.}\label{sil}
The silhouette feature encoder projects the individual frames $\{f_i\}_{i=1,...,m}$ of a gait sequence $G$ to their feature representings $\{v_{g,i}\}_{i=1,...,m}$, where $m$ is the number of frames. 
To verify the generality of using the body shape features to improve the gait recognition network, we choose four state-of-the-art gait recognition methods as the gait feature encoder for comparison:  GaitSet~\cite{chao2019gaitset}, GaitPart~\cite{fan2020gaitpart}, GaitGL~\cite{lin2021gaitgl}, and GLN~\cite{hou2020gln}.

\begin{figure*}[t]
    \centering
    \includegraphics[width=0.75\linewidth]{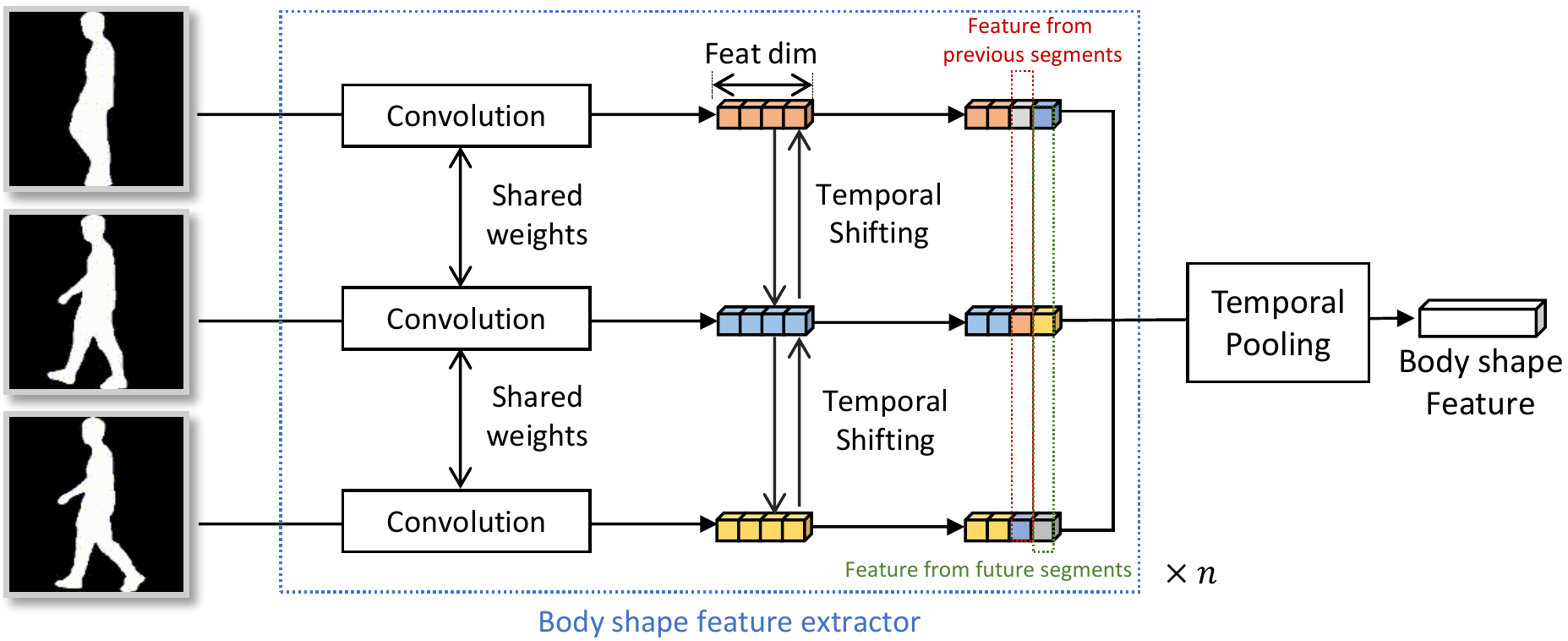}
    \caption{Our proposed body shape encoder for silhouette sequence input. $n$ represents the repeated time for the operations in the block.}
    \label{fig:TSnetwork}
\end{figure*}

\textbf{Body Shape Feature Encoder.}
To extract the body shape feature from the silhouette sequence, we input the gait sequence $G$ and project it to the feature space $v_{bs}$ representing the body shape of the person in the video. 
Extracting the body shape from a single gait sequence is difficult since the single binary silhouette only provides the boundary of a human body and lacks essential information. 
Thus we need the neighbor frames to help complete the missing information for extracting the whole human body shape. 
We show our proposed body shape feature encoder in Fig.~\ref{fig:TSnetwork}. 
The encoder consists of $n$ blocks, where every block includes a convolution and a temporal shifting (TS) operation. 
The convolution operator takes the frame-wise feature from the raw gait sequence or the previous layer as input, and operators in the same block share weight. 
After the convolution operation in each block, we follow \cite{lin2019tsm} to exchange $12.5\%$ of the features of part of the channels between the previous and future segments of video for temporal shifting. 

We preserve the first frame of the sequence's content, which should be exchanged with the previous frame since there is no frame before it. 
We also keep the feature from the future segments for the last frame. After the last layer of feature shifting, we do a temporal average pooling on the extracted feature sequences to generate the final body shape feature $v_{bs}$. 
With the features from two encoders, we concatenate the body shape feature $v_{bs}$ to each of the gait features $\{v_{g,i}\}_{i=1,...,m}$. 
We maxpool the features along the temporal and horizontal dimension following the implementation of \cite{chao2019gaitset,fan2020gaitpart,lin2021gaitgl,hou2020gln} and apply two fully-connected layers, whose dimensionalities match with the backbone network we used as the silhouette feature encoder, to generate feature $v_p$, representing the person's identity. 
We calculate the similarity between $v_p$ and the groundtruth $v_{gt}$ and calculate the identity loss $L_{ID}$ following \cite{chao2019gaitset,fan2020gaitpart,hou2020gln,lin2021gaitgl}.

\subsection{Human Body Prior Distillation and Transfer}
The purpose of inferring 3-D human body shapes is to separate movement patterns from variations in the appearance of 2-D silhouettes.
Due to the absence of ways to directly reconstruct the 3-D human body from the gait, we first extract the \HD{shape priors} from a small set of RGB frames in the gait sequence, then distill and transfer this information to the body shape feature encoder in the gait branch if corresponding RGB images are available. 

\textbf{Body Prior Inference.}
To infer 3-D body prior from the RGB images, we follow \cite{SMPL-X:2019} for using SMPL-X reconstructed from SMPLify-X as body shapes. 
Compared with other 3-D body reconstruction models such as PIFu~\cite{saito2019pifu} and PIFuHD~\cite{saito2020pifuhd}, SMPLify-X models human bodies with a strong human prior for the skinned body and outputs consistent results for the same person with different appearances, such as clothing, helping gait encoders to distinguish body shape from appearance variances in 2-D silhouettes. \HD{In addition, as a parametric method, SMPLify-X provides us the body shape feature decoupled from its pose and its strong prior can help us generate the complete shape even with some mild occlusions.}

Considering the time consumption for inferring body prior with SMPLify-X~\cite{SMPL-X:2019}, it is not feasible to extract the body prior for all frames in the video or image sequence.
Since the identity of the person within the same video is consistent, the inferred 3-D body prior without the pose should be identical across all frames in which the person is discernible.
Consequently, we infer the prior form based on one of the frames taken from the RGB sequence or video in conjunction with the gait sequence $G$.

To select this frame, we first extract the skeletons $\{s_i\}_{i=1,...,m}$ of the whole sequence using OpenPose~\cite{cao2019openpose}, followed by finding the longest sequence in the segments of $\{s_i\}$ with skeletons detectable and use the middle frame of this segment, which we annotate as $s_r$, to represent the body prior of the whole video. 
In this way, we can guarantee the quality of image $s_r$ used to infer the 3-D body prior since the longest segments with skeletons detectable can ensure stable and consistent performance for pose detection and estimation, making it easier for body prior extraction.

We then reconstruct the whole human body prior for $s_r$ by generating the shape feature $beta$ and 3-D pose $\theta$ following \cite{SMPL-X:2019}. 
$\beta$ is a 1-D vector with a size of 10 describing the appearance of the reconstructed body prior, and $\theta$ only includes 3-D joint positions. 
By combining the $\beta$ and $\theta$, we can reconstruct a full 3-D body model for a specific pose. 
In our experiments, we only use the $\beta$ as the body prior feature $v_{br}$ to guide the body shape based on silhouette $v_{bs}$. {No skeleton information is used for gait recognition to avoid the different accuracies of the prediction of skeleton.}

\textbf{Knowledge Distillation and Transfer.}\label{sec:transfer}
With the body prior features $v_{br}$ from the selected RGB frame and $v_{bs}$ from the silhouette sequence, we utilize $v_{br}$ to guide the training \HD{of feature $v_{bs}$ from the body shape feature encoder in the gait branch. }
We use CRD (Contrastive Representation Distillation)~\cite{tian2019contrastive} for distilling knowledge between features from $v_{br}$ and $v_{bs}$ following
\begin{equation}
\begin{split}
    L_{KD} = &\mathbb{E}_{q(v_{br} ,v_{bs}| C=1)}[\log\ h(v_{br}, v_{bs})] \\
    &+ \mathbb{E}_{q(v_{br} ,v_{bs}|C=0)}[\log(1-h(v_{br}, v_{bs}))]\\
    h(s, t) &= \frac{\exp(f_1(v_{bs})^T\cdot f_2(v_{br}))}{\exp(f_1(v_{bs})^T\cdot f_2(v_{br})) + \frac{N}{M}}
\end{split}
\end{equation}
where $f_1$ and $f_2$ are two linear projection layer with $L_2$ norm for projecting $v_{br}$ and $v_{bs}$. $N$ is the batch size and $M$ is the cardinality of the dataset. $C$ is 1 while $v_{br}$ and $v_{bs}$ are from the same identity and 0 if not. 
We will compare CRD with some other knowledge distillation methods in ablation studies.
During training, we have two different loss functions, $L_{ID}$ for gait recognition loss and $L_{KD}$ for knowledge distillation between the features of inferred 3-D body prior from the selected RGB frame and the gait sequence, $v_{br}$ and $v_{bs}$. We use a hyperparameter $\lambda$ for balancing two losses. The final objective is shown as
\begin{equation}\label{bala}
    L= \lambda_1 L_{ID} + \lambda_2 L_{KD}
\end{equation}
\HD{We set $\lambda_1$ to 1 empirically. Following ablations in the supplementary material, we set $\lambda_2$ to 1 for knowledge transfer for the examples with RGB images and 0 for others.}

\section{Experiments}

In this section, we show the details of our implementation for the experiment and the results.We first discuss our setups for the experiments in Sec.~\ref{sec:exp}, followed by our analysis based on the experiment results in Sec.~\ref{sec:res}.

\subsection{Experimental Setup}\label{sec:exp}
For experimental setup, we discuss datasets with the baseline methods and criteria. We also introduce the new setting of gait recognition where camera positions for training and test are mutually exclusive.

\begin{table*}[t]
\centering
\def\lw{0.8}
\def\la{1.5}
\def\ls{0.05}
\resizebox{.85\linewidth}{!}
{
\begin{tabular}{p{1.5cm}p{2.08cm}p{\ls cm}p{\lw cm}<{\centering}p{\lw cm}<{\centering}p{\lw cm}<{\centering}p{\lw cm}<{\centering}p{\lw cm}<{\centering}p{\lw cm}<{\centering}p{\lw cm}<{\centering}p{\lw cm}<{\centering}p{\lw cm}<{\centering}p{\lw cm}<{\centering}p{\lw cm}<{\centering}p{\ls cm}p{\lw cm}<{\centering}} \toprule
\multirow{2}{*}{Probe}&\multirow{2}{*}{Method} && \multicolumn{11}{c}{Camera Positions} && \multirow{2}{*}{Mean} \\

 \cline{4-14}  \\ [-8pt]
            &&& 0\degree   & 18\degree  & 36\degree  & 54\degree  & 72\degree  & 90\degree 
            &   108\degree & 126\degree & 144\degree & 162\degree & 180\degree && \\
\midrule
\multirow{8}{*}{NM \#5-6}
         & GaitSet \cite{chao2019gaitset}       && 91.1 & 98.0 & \textbf{99.6} & 97.8 & 95.4 & 93.8 & 95.7 & 97.5 & 98.1 & 97.0 & 88.2 && 95.6\\
         & GaitPart \cite{fan2020gaitpart}      && 94.0 & 98.7 & 99.3 & 98.8 & 94.8 & 92.6 & 96.4 & 98.3 & 99.0 & 97.4 & 91.2 && 96.4\\
         & GLN \cite{hou2020gln}                && 93.8 & 98.5 & 99.2 & 98.0 & 95.2 & 92.9 & 95.4 & 98.5 & 99.0 & 99.2 & 91.9 && 96.5\\
         & GaitGL \cite{lin2021gaitgl}          && 95.3 & 97.9 & 99.0 & 97.8 & 96.1 & 95.3 & 97.2 & \textbf{98.9} & 99.4 & 98.8 & \textbf{94.5} && 97.3\\
\cline{2-16}  \\ [-8pt]
         & GaitSet-HBS                          && 92.2 & 98.7 & 99.2 & 97.9 & 95.1 & 93.4 & 95.7 & 98.4 & 98.2 & 97.9 & 89.0 && 96.0\\
         & GaitPart-HBS                         && 93.2 & \textbf{98.9} & 99.4 & \textbf{98.9} & 95.1 & 91.9 & 96.5 & 98.8 & \textbf{99.5} & 98.4 & 91.7 && 96.6\\
         & GLN-HBS                              && 93.8 & 98.1 & 99.1 & 98.2 & 95.2 & 94.2 & 95.4 & 98.4 & 99.2 & \textbf{99.4} & 93.2 && 96.8\\
         & GaitGL-HBS                           && \textbf{96.0} & 98.3 & 99.2 & 97.8 & \textbf{96.4} & \textbf{95.9} & \textbf{97.4} & 98.7 & 99.2 & 98.7 & \textbf{94.5} && \textbf{97.5}\\
\midrule
\multirow{8}{*}{BG \#1-2}
         & GaitSet \cite{chao2019gaitset}       && 87.0 & 93.8 & 94.6 & 92.9 & 88.2 & 83.0 & 86.6 & 92.6 & 95.7 & 92.9 & 83.4 && 90.1\\
         & GaitPart \cite{fan2020gaitpart}      && 89.5 & 94.5 & 95.3 & 93.5 & 88.5 & 83.9 & 89.0 & 93.6 & 96.0 & 94.1 & 85.3 && 91.2\\
         & GLN \cite{hou2020gln}                && 92.2 & 95.6 & 96.7 & 94.3 & 91.8 & 87.8 & 91.4 & 95.1 & 96.3 & 95.7 & 87.2 && 93.1\\
         & GaitGL \cite{lin2021gaitgl}          && \textbf{93.0} & 95.7 & 97.0 & \textbf{95.9} & 93.3 & \textbf{90.0} & 91.9 & 96.8 & 97.5 & 96.9 & 90.7 && 94.4\\
\cline{2-16}  \\ [-8pt]
         & GaitSet-HBS                          && 89.7 & 93.8 & 95.9 & 93.3 & 87.1 & 83.1 & 87.4 & 91.9 & 94.1 & 93.7 & 85.1 && 90.5\\
         & GaitPart-HBS                         && 90.1 & 93.6 & 95.7 & 94.4 & 89.9 & 85.8 & 89.9 & 94.0 & 96.0 & 92.7 & 86.4 && 91.7\\
         & GLN-HBS                              && 91.7 & \textbf{96.6} & 96.6 & 95.2 & 90.9 & 88.1 & 91.5 & 95.4 & 96.6 & 96.8 & 89.8 && 93.6\\
         & GaitGL-HBS                           && \textbf{93.0} & 96.0 & \textbf{97.3} & \textbf{95.9} & \textbf{93.7} & 89.5 & \textbf{92.9} & \textbf{97.0} & \textbf{98.3} & \textbf{97.4} & \textbf{92.2} && \textbf{94.8}\\
\midrule
\multirow{8}{*}{CL \#1-2}
         & GaitSet \cite{chao2019gaitset}       && 71.0 & 82.6 & 84.0 & 80.0 & 71.7 & 69.1 & 72.1 & 76.7 & 78.5 & 77.2 & 63.4 && 75.1\\
         & GaitPart \cite{fan2020gaitpart}      && 72.5 & 82.8 & 86.0 & 82.2 & 79.5 & 71.0 & 77.7 & 80.8 & 82.9 & 81.4 & 67.7 && 78.6\\
         & GLN \cite{hou2020gln}                && 78.5 & 90.4 & 90.3 & 85.1 & 80.2 & 75.8 & 78.1 & 81.8 & 80.9 & 83.2 & \textbf{72.6} && 81.5\\
         & GaitGL \cite{lin2021gaitgl}          && 71.7 & \textbf{90.5} & \textbf{92.4} & 89.4 & \textbf{84.9} & \textbf{78.1} & 83.1 & \textbf{87.5} & 89.1 & 83.9 & 67.4 && 83.5\\
\cline{2-16}  \\ [-8pt]
         & GaitSet-HBS                          && 72.9 & 84.1 & 83.7 & 79.6 & 73.0 & 70.5 & 73.1 & 76.6 & 79.8 & 78.3 & 64.6 && 76.0\\
         & GaitPart-HBS                         && 75.9 & 84.8 & 86.5 & 84.6 & 77.4 & 74.4 & 78.6 & 82.4 & 83.5 & 80.5 & 67.6 && 79.7\\
         & GLN-HBS                              && 77.7 & 89.4 & 91.9 & 87.0 & 84.1 & \textbf{78.1} & 81.6 & 83.8 & 85.2 & 83.8 & \textbf{72.6} && 83.2\\
         & GaitGL-HBS                           && \textbf{75.8} & \textbf{90.5} & 92.3 & \textbf{90.0} & 84.0 & 77.9 & \textbf{83.3} & 87.3 & \textbf{89.3} & \textbf{85.1} & 69.8 && \textbf{84.1} \\
\bottomrule
\end{tabular}
}
\medskip
\caption{Gait recognition results on CASIA-B dataset, excluding identical-view cases.} 
\label{tab:casiab-1}
\end{table*}

\textbf{Datasets.}
We conduct our experiments on two public datasets,  CASIA-B~\cite{yu2006framework} and OU-MVLP~\cite{takemura2018multi}.

\textit{CASIA-B}~\cite{yu2006framework} is a gait recognition dataset with 124 objects with 10 different walking variants for each subject, where 6 are for normal walking (NM), 2 for walking while carrying bags (BG) and 2 for different clothing (CL). 
Each variant is recorded from 11 different camera viewpoints between 0\degree\ and 180\degree\ with 18 as the gap, making 110 videos for each subject. We follow \cite{chao2019gaitset,fan2020gaitpart,hou2020gln,lin2021gaitgl} to use silhouette sequences of the first 74 subjects for training. 
{During inference, we use the first four walking variances in normal walking conditions (NM) as the gallery set, which is the identity library for the test set.
The remaining 2 variants in NM, along with the sequences in BG and CL for the remaining 50 subjects, are used as probes for evaluation to find the correct identity in the gallery set.}

In addition to using all the camera positions for supervised gait recognition, we introduce a new zero-shot setting where camera viewpoints used for training and testing are mutually exclusive. For all the camera viewpoints in the dataset, we only sample partial angles between 0\degree\ and 90\degree\ for training and use the viewpoints between 108\degree\ and 180\degree\ for inference to assess the model's performance when encountering novel viewpoints. {We will further discuss about this dataset in the supplementary material.}

\textit{OUMVLP}~\cite{takemura2018multi} is a large gait recognition dataset with 10,307 subjects. Each subject has 2 different sequences for normal walking (NM) recorded from 14 different camera positions, resulting in 28 gait sequences for each subject. The camera viewpoints are evenly distributed from 0\degree\ to 90\degree\ and 180\degree\ and 270\degree, with a 15-degree gap. Following \cite{chao2019gaitset,fan2020gaitpart,hou2020gln,lin2021gaitgl}, we use the 5,153 subjects with an odd index between the 1-\textit{st} and 10,305-\textit{th} as training examples and the remaining 5,154 for inference, where the first sequence for each subject is the gallery set and the second as the probe.

\textbf{Implementation Details.}
To extract the silhouette features, we follow the original settings of baseline methods \cite{chao2019gaitset,fan2020gaitpart,lin2021gaitgl,hou2020gln} for setting the hyperparameters for the model. For GaitPart~\cite{fan2020gaitpart}, GaitSet~\cite{chao2019gaitset} and GaitGL~\cite{lin2021gaitgl}, we resize input gait sequence $g$ to the size of $64\times 44$. We use Adam optimizer with 1e-4 as the learning rate and 0.9 as the momentum. We set the margin in separate triplet loss as 0.2. Batch size is set to  (8, 16) for CASIA-B, and (32, 16) for OUMVLP. We set the maximum iteration and weight decay following \cite{chao2019gaitset,fan2020gaitpart,lin2021gaitgl,hou2020gln}.
For GLN~\cite{hou2020gln}, the initial gait sequence is sampled to $128\times 88$. We use SGD with 0.1 as the initial learning rate and reduce it to $\frac{1}{10}$ three times during training. The learning rate is reduced every 10,000 steps for CASIA-B and every 50,000 steps for OUMVLP.

\begin{table*}[tb!]
\centering
\def\lw{2}
\def\ls{0.05}
\resizebox{0.76\linewidth}{!}
{
\begin{tabular}{p{1.5cm}p{1.5cm}<{\centering}p{0.7cm}<{\centering}p{\ls cm}p{\lw cm}<{\centering}p{\lw cm}<{\centering}p{\lw cm}<{\centering}p{\lw cm}<{\centering}p{\ls cm}p{1.5 cm}<{\centering}} 
\toprule
\multirow{2}{*}{Probe}  & \multirow{2}{*}{Stats} & \multirow{2}{*}{HBS} && \multicolumn{4}{c}{Method} && Avg.\\
\cline{5-9}  \\ [-8pt]
& & && GaitSet \cite{chao2019gaitset} & GaitPart \cite{fan2020gaitpart} & GLN \cite{hou2020gln} & GaitGL \cite{lin2021gaitgl} && Change  \\ 
\midrule
\multirow{6}{*}{NM \#5-6} & \multirow{3}{*}{Mean ($\uparrow$)} & \xmark && 95.6 & 96.4 & 96.5 & 97.3  \\
& &\cmark && 96.0 & 96.6 & 96.8 & 97.5 && \textit{\textbf{+0.3}}\\
\cline{3-8}  \\ [-8pt]
& & $\Delta$ && \textit{+0.3} & \textit{+0.2} & \textit{+0.3} & \textit{+0.2} \\
\cline{2-10}  \\ [-8pt]
 & \multirow{3}{*}{STD. ($\downarrow$)} & \xmark && 3.4 & 2.8 & 2.8 & 1.7 & \\
& &\cmark && 3.2 & 3.0 & 2.4 & 1.6 && \textit{\textbf{-0.1}} \\
\cline{3-8}  \\ [-8pt]
& & $\Delta$ && \textit{-0.2} & \textit{+0.2} & \textit{-0.4} & \textit{-0.1} \\
\midrule
\multirow{6}{*}{BG \#1-2}& \multirow{3}{*}{Mean ($\uparrow$)} 
  & \xmark && 90.1 & 91.2 & 93.1 & 94.4 & \\
& & \cmark && 90.5 & 91.7 & 93.6 & 94.8  && \textit{\textbf{+0.4}}\\
\cline{3-8}  \\ [-8pt]
& & $\Delta$ && \textit{+0.4} & \textit{+0.5} & \textit{+0.5} & \textit{+0.4} \\
\cline{2-10}  \\ [-8pt]
 & \multirow{3}{*}{STD. ($\downarrow$)} 
  & \xmark && 4.6 & 4.2 & 3.3 & 2.7 & \\
& & \cmark && 4.2 & 3.5 & 3.2 & 2.7 && \textit{\textbf{-0.3}}\\
\cline{3-8}  \\ [-8pt]
& & $\Delta$ && \textit{-0.4} & \textit{-0.7} & \textit{-0.1} & \textit{0.0} \\
\midrule
\multirow{6}{*}{CL \#1-2}& \multirow{3}{*}{Mean ($\uparrow$)} 
  & \xmark && 75.1 & 78.6 & 81.5 & 83.5 \\
& & \cmark && 76.0 & 79.7 & 83.2 & 84.1 && \textit{\textbf{+1.1}}\\
\cline{3-8}  \\ [-8pt]
& & $\Delta$ && \textit{+0.9} & \textit{+1.1} & \textit{+1.7} & \textit{+0.6} \\
\cline{2-10}  \\ [-8pt]
 & \multirow{3}{*}{STD. ($\downarrow$)} & \xmark && 6.2 & 5.8 & 5.5 & 8.0 & \\
& &\cmark && 5.9 & 5.6 & 5.5 & 7.0 && \textit{\textbf{-0.4}}\\
\cline{3-8}  \\ [-8pt]
& & $\Delta$ && \textit{-0.3} & \textit{-0.2} & \textit{0.0} & \textit{-1.0} \\
\bottomrule
\end{tabular}
}\medskip
\caption{Statistics analysis for supervised results on CASIA-B dataset, excluding identical-view cases. ($\uparrow$) indicates that larger values show better performance, while ($\downarrow$) indicates that lower values are better. \HD{$\Delta$ indicates the change between the method with and without HBS.}} 
\label{tab:casiab-2} 
\end{table*}

\begin{table*}[t!]
\centering
\def\lw{0.8}
\def\ls{0.05}
\resizebox{.9\linewidth}{!}
{
\begin{tabular}{p{1.5cm}p{2.08cm}p{\ls cm}p{\lw cm}<{\centering}p{\lw cm}<{\centering}p{\lw cm}<{\centering}p{\lw cm}<{\centering}p{\lw cm}<{\centering}p{\lw cm}<{\centering}p{\ls cm}p{\lw cm}<{\centering}p{\lw cm}<{\centering}p{\lw cm}<{\centering}p{\lw cm}<{\centering}p{\lw cm}<{\centering}p{\ls cm}p{\lw cm}<{\centering}p{\ls cm}p{\lw cm}<{\centering}} \toprule
\multirow{2}{*}{Probe}&\multirow{2}{*}{Method} && \multicolumn{6}{c}{Training Viewpionts} && \multicolumn{5}{c}{Test Viewpionts} && \multirow{2}{*}{Mean} && Avg. \\

 \cline{4-9} \cline{11-15}  \\ [-8pt]
            &&& 0\degree   & 18\degree  & 36\degree  & 54\degree  & 72\degree  & 90\degree 
            &&  108\degree & 126\degree & 144\degree & 162\degree & 180\degree &&  && Diff. \\
\midrule
\multirow{16}{*}{NM \#5-6}
         & GLN                                  && \multicolumn{6}{c}{(All Camera Positions)}            && 95.4 & 98.5 & 99.0 & 99.2 & 91.9 && 96.8 &&\multirow{2}{*}{\textit{+0.3}}\\
         & GLN-HBS                              && \multicolumn{6}{c}{(All Camera Positions)}            && 95.4 & 98.4 & 99.2 & 99.4 & 93.1 && 97.1\\
\cline{2-19}  \\ [-8pt]   
         & GLN                                  && \cmark & \cmark & \cmark & \cmark & \cmark & \cmark && 82.3 & 91.8 & 95.8 & 89.3 & 78.0 && 87.4\\
         & GLN                                  && \cmark &        & \cmark &        & \cmark &        && 79.7 & 88.2 & 94.7 & 87.5 & 78.0 && 85.7\\
         & GLN                                  && \cmark &        &        & \cmark &        &        && 74.0 & 89.3 & 93.5 & 83.8 & 76.3 && 83.3 && \multirow{2}{*}{\textit{+1.3}}\\
\cline{2-17}  \\ [-8pt]
         & GLN-HBS                              && \cmark & \cmark & \cmark & \cmark & \cmark & \cmark && 85.5 & 93.5 & 97.3 & 89.0 & 82.3 && 89.5\\
         & GLN-HBS                              && \cmark &        & \cmark &        & \cmark &        && 82.0 & 88.5 & 93.5 & 91.8 & 77.5 && 86.7\\
         & GLN-HBS                              && \cmark &        &        & \cmark &        &        && 77.2 & 89.5 & 94.5 & 83.2 & 76.5 && 84.2\\
\cline{2-19}  \\ [-8pt]         
         & GaitGL                               && \multicolumn{6}{c}{(All Camera Positions)}            && 97.2 & 98.9 & 99.4 & 98.8 & 94.5 && 97.8 &&\multirow{2}{*}{\textit{-0.1}}\\
         & GaitGL-HBS                           && \multicolumn{6}{c}{(All Camera Positions)}            && 97.4 & 98.7 & 99.2 & 98.7 & 94.5 && 97.7\\
\cline{2-19}  \\ [-8pt]   
         & GaitGL                               && \cmark & \cmark & \cmark & \cmark & \cmark & \cmark && 84.5 & 93.3 & 95.8 & 92.3 & 75.0 && 88.2\\
         & GaitGL                               && \cmark &        & \cmark &        & \cmark &        && 81.5 & 90.0 & 92.8 & 89.5 & 69.5 && 84.7\\
         & GaitGL                               && \cmark &        &        & \cmark &        &        && 76.3 & 91.0 & 91.3 & 86.2 & 69.7 && 82.9 && \multirow{2}{*}{\textit{+1.4}}\\
\cline{2-17}  \\ [-8pt]                  
         & GaitGL-HBS                           && \cmark & \cmark & \cmark & \cmark & \cmark & \cmark && 88.0 & 93.8 & 95.5 & 92.5 & 78.8 && 89.7\\
         & GaitGL-HBS                           && \cmark &        & \cmark &        & \cmark &        && 82.8 & 90.5 & 93.0 & 89.7 & 71.7 && 85.6\\
         & GaitGL-HBS                           && \cmark &        &        & \cmark &        &        && 79.0 & 92.0 & 91.7 & 88.5 & 71.5 && 84.6\\
\midrule
\multirow{16}{*}{BG \#1-2}
         & GLN                                  && \multicolumn{6}{c}{(All Camera Positions)}            && 91.4 & 95.1 & 96.3 & 95.7 & 87.2 && 93.1 &&\multirow{2}{*}{\textit{+0.9}}\\
         & GLN-HBS                              && \multicolumn{6}{c}{(All Camera Positions)}            && 91.5 & 95.4 & 96.6 & 96.8 & 89.8 && 94.0\\
\cline{2-19}  \\ [-8pt]   
         & GLN                                  && \cmark & \cmark & \cmark & \cmark & \cmark & \cmark && 72.0 & 83.0 & 87.3 & 80.1 & 75.0 && 79.5\\
         & GLN                                  && \cmark &        & \cmark &        & \cmark &        && 70.7 & 79.2 & 88.5 & 80.0 & 73.5 && 78.4\\
         & GLN                                  && \cmark &        &        & \cmark &        &        && 65.0 & 81.5 & 86.5 & 79.6 & 65.5 && 75.6 && \multirow{2}{*}{\textit{+1.3}}\\
\cline{2-17}  \\ [-8pt]
         & GLN-HBS                              && \cmark & \cmark & \cmark & \cmark & \cmark & \cmark && 74.5 & 85.0 & 88.8 & 82.1 & 74.0 && 80.9\\
         & GLN-HBS                              && \cmark &        & \cmark &        & \cmark &        && 73.2 & 82.0 & 88.3 & 86.4 & 72.7 && 80.5\\
         & GLN-HBS                              && \cmark &        &        & \cmark &        &        && 69.5 & 81.5 & 86.5 & 77.3 & 65.2 && 76.0\\
\cline{2-19}  \\ [-8pt]         
         & GaitGL                               && \multicolumn{6}{c}{(All Camera Positions)}            && 91.9 & 96.8 & 97.5 & 96.9 & 90.7 && 94.8
         &&\multirow{2}{*}{\textit{+0.8}}\\
         & GaitGL-HBS                           && \multicolumn{6}{c}{(All Camera Positions)}            && 92.9 & 97.0 & 98.3 & 97.4 & 92.2 && 95.6\\
\cline{2-19}  \\ [-8pt]   
         & GaitGL                               && \cmark & \cmark & \cmark & \cmark & \cmark & \cmark && 74.3 & 83.8 & 90.0 & 88.1 & 69.3 && 81.1\\
         & GaitGL                               && \cmark &        & \cmark &        & \cmark &        && 72.2 & 81.0 & 85.7 & 84.1 & 61.5 && 76.9\\
         & GaitGL                               && \cmark &        &        & \cmark &        &        && 64.7 & 82.7 & 86.5 & 78.5 & 66.8 && 75.9 && \multirow{2}{*}{\textit{+1.6}}\\
\cline{2-17}  \\ [-8pt]                  
         & GaitGL-HBS                           && \cmark & \cmark & \cmark & \cmark & \cmark & \cmark && 75.5 & 87.8 & 91.8 & 87.4 & 70.5 && 82.6\\
         & GaitGL-HBS                           && \cmark &        & \cmark &        & \cmark &        && 70.2 & 81.2 & 89.3 & 84.6 & 66.3 && 78.3\\
         & GaitGL-HBS                           && \cmark &        &        & \cmark &        &        && 70.8 & 83.2 & 87.2 & 80.8 & 67.0 && 77.8\\
\midrule
\multirow{16}{*}{CL \#1-2}
         & GLN                                  && \multicolumn{6}{c}{(All Camera Positions)}            && 78.1 & 81.8 & 80.9 & 83.2 & 72.6 && 79.3 &&\multirow{2}{*}{\textit{+2.1}}\\
         & GLN-HBS                              && \multicolumn{6}{c}{(All Camera Positions)}            && 81.6 & 83.8 & 85.2 & 83.8 & 72.6 && 81.4\\
\cline{2-19}  \\ [-8pt]   
         & GLN                                  && \cmark & \cmark & \cmark & \cmark & \cmark & \cmark && 57.3 & 60.0 & 67.0 & 56.0 & 46.3 && 57.3\\
         & GLN                                  && \cmark &        & \cmark &        & \cmark &        && 50.0 & 62.5 & 67.5 & 58.5 & 44.8 && 56.5\\
         & GLN                                  && \cmark &        &        & \cmark &        &        && 45.0 & 54.7 & 59.3 & 52.0 & 44.5 && 51.1 && \multirow{2}{*}{\textit{+2.3}}\\
\cline{2-17}  \\ [-8pt]
         & GLN-HBS                              && \cmark & \cmark & \cmark & \cmark & \cmark & \cmark && 57.8 & 62.5 & 68.3 & 61.5 & 46.8 && 59.4\\
         & GLN-HBS                              && \cmark &        & \cmark &        & \cmark &        && 54.8 & 62.5 & 66.5 & 62.7 & 44.3 && 58.2\\
         & GLN-HBS                              && \cmark &        &        & \cmark &        &        && 47.5 & 58.0 & 64.0 & 55.3 & 45.5 && 54.1\\
\cline{2-19}  \\ [-8pt]           
         & GaitGL                               && \multicolumn{6}{c}{(All Camera Positions)}            && 83.1 & 87.5 & 89.1 & 83.9 & 67.4 && 82.2 &&\multirow{2}{*}{\textit{+0.8}}\\
         & GaitGL-HBS                           && \multicolumn{6}{c}{(All Camera Positions)}            && 83.3 & 87.3 & 89.3 & 85.1 & 69.8 && 83.0\\
\cline{2-19}  \\ [-8pt]     
         & GaitGL                               && \cmark & \cmark & \cmark & \cmark & \cmark & \cmark && 58.8 & 68.5 & 73.3 & 66.8 & 44.0 && 62.3\\
         & GaitGL                               && \cmark &        & \cmark &        & \cmark &        && 53.2 & 63.7 & 71.2 & 63.5 & 41.0 && 58.5\\
         & GaitGL                               && \cmark &        &        & \cmark &        &        && 48.5 & 62.0 & 64.2 & 51.5 & 43.3 && 53.9 && \multirow{2}{*}{\textit{+1.8}}\\
\cline{2-17}  \\ [-8pt]                  
         & GaitGL-HBS                           && \cmark & \cmark & \cmark & \cmark & \cmark & \cmark && 61.5 & 70.8 & 76.0 & 72.3 & 47.3 && 65.6\\
         & GaitGL-HBS                           && \cmark &        & \cmark &        & \cmark &        && 59.8 & 63.5 & 71.8 & 60.5 & 44.0 && 59.9\\
         & GaitGL-HBS                           && \cmark &        &        & \cmark &        &        && 49.0 & 62.3 & 69.8 & 51.0 & 41.3 && 54.7\\
\bottomrule
\end{tabular}
}
\medskip
\caption{Gait recognition results for novel camera viewpoints on CASIA-B dataset. Viewpoints used for the training and inference stages are mutually exclusive. Supervised results, where all viewpoints are available for training, are shown at the top of each set.} 
\label{tab:casiab-part}
\end{table*}
For the body shape encoder at the gait branch, we apply the temporal shifting modules to MobileNet-v2 \cite{sandler2018mobilenetv2} and set $n$ to 6  following \cite{lin2019tsm} with the same learning rates and hyperparameters as the silhouette feature extraction model. We use CRD as our knowledge distillation method, where the ablation studies for other methods can be found in the ablation studies in the supplementary material. 

To fuse the inferred body shape feature with the gait features from the silhouette feature encoder, we append the features before the last fully-connected layers for each backbone, since the features before the temporal or set pooling are the high-level feature representing the frame-level identity, and the inferred 3-D body shape representation can give additional guidance for identity encoding. For GaitPart, we append the 3-D body shape feature to all the part features to help each feature for a specific part understand the global body shape along with its local patterns. The input of fully-connected layers is set to the original size of the identity feature plus 10 (size of $v_{bs}$) for each model \cite{chao2019gaitset,fan2020gaitpart,lin2021gaitgl,hou2020gln} after feature concatenation. 

{\textbf{RGB Data for Knowledge Distillation.} For 3-D human body prior extraction, we use the latent feature $\beta$ in SMPL-X \cite{loper2015smpl} model and normalize features in training set to $(0, 0.1)$ gaussian distribution.  To supervise the generation of body shape feature in the gait branch, we select $20\%$ of sequences in the CASIA-B sequence for the data distillation and transfer. Since OUMVLP does not provide the RGB video sequences, we apply the body shape feature encoder for the gait branch pretrained on the CASIA-B subset and keep it frozen during training for feature extraction for all the examples in the OUMVLP dataset. 
}

\textbf{Details for Identity Loss $L_{ID}$.}
For the selection of identity loss function $L_{ID}$, we follow the implementation of each baseline method \cite{chao2019gaitset,fan2020gaitpart,hou2020gln,lin2021gaitgl}. For GaitSet-HBS, GaitPart-HBS and GLN-HBS, We use the triplet loss with its margin set to 0.2 as $L_{ID}$. For GaitGL-HBS, in addition to the triplet loss with the same margin, we use a cross-entropy loss for predicting the identity, which is represented as a one-hot vectors; weights for both losses are set to 1.

\begin{table*}[tb]
\centering
\def\lw{0.9}
\def\ls{0.06}
\resizebox{.98\linewidth}{!}
{
\begin{tabular}{p{2.08cm}p{\ls cm}p{\lw cm}<{\centering}p{\lw cm}<{\centering}p{\lw cm}<{\centering}p{\lw cm}<{\centering}p{\lw cm}<{\centering}p{\lw cm}<{\centering}p{\lw cm}<{\centering}p{\lw cm}<{\centering}p{\lw cm}<{\centering}p{\lw cm}<{\centering}p{\lw cm}<{\centering}p{\lw cm}<{\centering}p{\lw cm}<{\centering}p{\lw cm}<{\centering}p{\ls cm}p{0.9 cm}<{\centering}} \toprule
\multirow{2}{*}{Method} && \multicolumn{14}{c}{Camera Positions} && \multirow{2}{*}{Mean} \\

 \cline{3-16}  \\ [-8pt]
            && 0\degree   & 15\degree  & 30\degree  & 45\degree  & 60\degree  & 75\degree  & 90\degree 
            &   180\degree & 195\degree & 210\degree & 225\degree & 240\degree & 255\degree & 270\degree && \\
\midrule
GEINet \cite{shiraga2016geinet} && 23.2 & 38.1 & 48.0 & 51.8 & 47.5 & 48.1 & 43.8 & 27.3 & 37.9 & 46.8 & 49.9 & 45.9 & 45.7 & 41.0 && 42.5 \\
GaitSet \cite{chao2019gaitset} && 79.2 & 87.7 & 89.9 & 90.1 & 87.9 & 88.6 & 87.7 & 81.7 & 86.4 & 89.0 & 89.2 & 87.2 & 87.7 & 86.2 && 87.0 \\
GaitPart \cite{fan2020gaitpart} &&  82.8 & 89.2 & 90.9 & 91.0 & 89.7 & 89.9 & 89.3 & 85.1 & 87.7 & 90.0 & 90.1 & 89.0 & 89.0 & 88.1 && 88.7 \\
GaitGL \cite{lin2021gaitgl} && 84.2 & 89.8 & 91.3 & \textbf{91.7} & 90.8 & \textbf{91.0} & 90.4 & 88.1 & 88.2 & \textbf{90.5} & \textbf{90.5} & 89.5 & \textbf{89.7} & 88.8 && 89.6 \\
\midrule
GaitSet-HBS && 79.0 & 87.9 & 90.4 & 90.6 & 88.4 & 89.2 & 88.4 & 82.3 & 87.1 & 89.6 & 89.6 & 87.7 & 88.4 & 86.9 && 87.5\\
GaitPart-HBS && 82.4 & 89.1 & 91.1 & 91.3 & 89.8 & 90.2 & 89.7 & 84.8 & 88.0 & 90.3 & 90.3 & 89.2 & 89.4 & 88.4 && 88.9\\
\textcolor{black}{GaitGL-HBS}  && \textbf{84.7} & \textbf{90.2} & \textbf{91.4} & \textbf{91.7} & \textbf{90.9} & \textbf{91.0} & \textbf{90.5} & \textbf{88.4} & \textbf{88.7} & \textbf{90.5} & \textbf{90.5} & \textbf{89.6} & 89.6 & \textbf{88.9} && \textbf{89.8}\\
\bottomrule
\end{tabular}
}
\medskip
\caption{Gait recognition results on OUMVLP dataset, excluding identical-view cases. } 
\label{tab:oumvlp}
\end{table*}
\textbf{Baseline Methods.}
Since our method is an additional to the existing gait recognition methods, we compare with four state-of-the-art deep-learning gait recognition methods: GaitSet~\cite{chao2019gaitset}, GaitPart~\cite{fan2020gaitpart}, GaitGL~\cite{lin2021gaitgl} and GLN~\cite{hou2020gln}.  We compare the baseline methods with and without inferred 3-D human body shape on both datasets. For ablation studies, we conduct our experiments on GaitGL~\cite{lin2021gaitgl} and GLN~\cite{hou2020gln}, since these are the two state-of-the-art methods for gait recognition. We exclude \HD{GaitView \cite{chai2021silhouette} and Gait3D \cite{zheng2022gait} as they have extra supervision or additional input modality (framewise skeletons and body meshes) from RGB images. 
We also exclude earlier methods, such as \cite{wu2016comprehensive,shiraga2016geinet,song2019gaitnet}, which not show state-of-the-art performance.}

\textbf{Inference and Metrics.} 
We assess $L_2$ similarity between features extracted from examples from gallery and probe sets, excluding the identical-view cases. We calculate the top-1 accuracies for finding the response with the smallest $L_2$ distance among the examples in the gallery to each example in the probe set.

\subsection{Results and Analysis}\label{sec:res}

In this subsection, we present the results and analysis on CASIA-B~\cite{yu2006framework} and OUMVLP~\cite{takemura2018multi}. We further conduct ablations on CASIA-B for the selection of 3-D body shape features along with knowledge distillation and transfer.

\textbf{Results on CASIA-B.}
We show the results for CASIA-B in Table~\ref{tab:casiab-1}. Methods ending with `HBS', which is the abbriviation of \textbf{H}uman \textbf{B}ody \textbf{S}hape, are the ones with inferred 3-D human body features compared with baseline methods. 
In addition, we summarize the statistics for the performance on CASIA-B in Table~\ref{tab:casiab-2}, where we compare the models with and without features for the inferred human body. 
Mean and STD values in Table~\ref{tab:casiab-2} refer to the average and standard deviation values of performance for different viewpoints for the same model. We have the following observations:

\begin{enumerate}
    \setlength{\itemsep}{0pt}
    \item \textbf{Better performance.} Table~\ref{tab:casiab-2} shows that the models with inferred human body shapes outperform the original ones on all four baselines for all three splits. For most of the viewpoints shown, the best performances among all models also appear in the model with the inferred 3-D body shape. With the knowledge of the boundary of the skinned human body model, gait recognition models are capable of focusing on the motions instead of the appearances in 2-D silhouettes.
    
    \item \textbf{Stability at different viewpoints.} In addition to the average performance for all camera viewpoints, we observe the standard deviations for the accuracies at different viewpoints reduce after using inferred human body shapes. Even for those models with no improvement on the mean value, e.g., GaitPart-HBS compared with GaitPart on the NM split, the standard deviation still reduces. With the inferred 3-D body shape, consistent for all camera positions, models can show additional robustness to the camera viewpoints and have more stable performances.
    \item \textbf{Different appearance variances.} BG and CL sets have higher average accuracy than NM, whose gait appearances are similar. In BG and CL sets, the silhouette sequence individual is carrying different bags or wearing different outfits, affecting the binarized silhouette. Focusing on appearance differences hurts the gait recognition model. Since inferred 3-D human body shapes are skinned models, they are stable and resilient to these fluctuations. Gait recognition models may detect the consistent body shapes and reduce non-human body content, exhibiting benefits.
\end{enumerate}

\textbf{Zero-shot Results for Novel Viewpoints.} 
In addition to the results on existing viewpoints, we assess the model on the novel viewpoints on CASIA-B dataset in Table~\ref{tab:casiab-part} with GaitGL and GLN, the two of the best performing baselines. Instead of using silhouette sequence from all the viewpoints for both training and inference, we only use part of the viewpoints for training, and viewpoints used for training and inference are mutually exclusive. We notice that when gait recognition models encounter novel viewpoints not seen before, using the inferred human body shape gives a consistent improvement compared with the baseline methods. Although these novel camera positions are unavailable during training, the consistency of the 3-D human body shape helps gait recognition models extract motion information from a new camera position for identification.

We further reduce the number of available viewpoints during training to assess the robustness of gait recognition models learning from fewer examples. With fewer viewpoints available in the training set, performances for all the methods are decreasing. However, GaitGL~\cite{lin2021gaitgl} and GLN~\cite{hou2020gln} with inferred 3-D human body shape still show a consistent improvement compared to the model without body shapes, showing the 3-D body shape can give consistent guidance at different amounts of data.

\textbf{Results on OUMVLP.}
We show the results for the OUMVLP dataset in Table~\ref{tab:oumvlp}. Since the OUMVLP dataset does not provide the original RGB frames, we apply the knowledge distillation model pretrained on the training set of CASIA-B to infer human body shape directly from the silhouette sequences. 
Compared to baseline methods, inferring 3-D body shape for gait recognition consistently outperforms original methods, showing good generalization ability and robustness of body shape feature encoders across different datasets. 
{Examples in OUMVLP are all normal walking with fewer variations, which explains the limited improvement as NM sets for CASIA-B.}

\section{Conclusion}
In this paper, we propose the exploitation of inferring 3-D body shape from gait sequence to disentangle gait motion from appearance variances of 2-D images. In addition to the gait pattern analysis, we distill the 3-D body shape features from selected RGB frames and transfer them to gait sequences via feature exchanging between neighbor frames. We assess our method with four state-of-the-art gait recognition methods and show better results on two public datasets at both seen and novel camera viewpoints.
\subsection*{Acknowledgement}
This research is based upon work supported in part by the Office of the Director of National Intelligence (ODNI), Intelligence Advanced Research Projects Activity (IARPA), via [2022-21102100007]. The views and conclusions contained herein are those of the authors and should not be interpreted as necessarily representing the official policies, either expressed or implied, of ODNI, IARPA, or the U.S. Government. The U.S. Government is authorized to reproduce and distribute reprints for governmental purposes notwithstanding any copyright annotation therein.

{\small
\bibliographystyle{ieee_fullname}
\bibliography{egbib}
}

\appendix
\null
\begin{table*}
\vskip .375in
\begin{center}
  {\Large \bf \papertitle \\\textit{Supplementary Material} \par}
\end{center}
\end{table*}

\newpage
In this supplementary document, we present some further experimental details and results that could not fit in the main paper. We discuss the motivation and details for the new setting for the CASIA-B dataset with novel camera viewpoints as further experiment details, followed by some experimental details and additional ablation studies for hyperparameters we choose in the main paper; these include the balancing term $\lambda$ in the final loss function and the ratio of feature exchange in the temporal shift operation. We then show some visualization results for the inferred body shapes directly from silhouette compared with the reconstruction results by SMPLify-X \cite{SMPL-X:2019} for selected RGB frames.

\section{Experiment Details}

{
\textbf{Discussion for the novel view settings.} In addition to the original CASIA-B setting in which the training and test set share the same viewpoints, the new setting of CASIA-B only includes 2 to 6 viewpoints in the training set, while we evaluate the model on the test viewpoints of the remaining five camera viewpoints, 108\degree, 126\degree, 144\degree, 162\degree, and 180\degree. In a real-world instance of silhouettes taken by a camera, the camera's perspective can come from any direction, which is the primary purpose of introducing this new setting. Compared to the original setting, our setting is more suitable for evaluating the generalization capacity of the gait recognition model when meeting novel camera viewpoints.}

\textbf{Variations for Silhouette Feature Encoder.}
In the experiment, we choose four methods as our sihouette feature encoder: GaitSet, GaitPart, GaitGL and GLN.
\textit{GaitSet}~\cite{chao2019gaitset} uses the frame sequence in the gait video as a \textit{set} of independent frames. By using set processing methods, such as set pooling, GaitSet can extract set-level features for preserving spatial and temporal information. 
\textit{GaitPart}~\cite{fan2020gaitpart} introduces split the gait image into four different parts and assess the motion pattern for each part separately to focus on more local movements. 
\textit{GLN}~\cite{hou2020gln} learns both discriminative and compact representations from the silhouettes. It extracts both silhouette-level and set-level features from different stages for gait recognition. 
\textit{GaitGL}~\cite{lin2021gaitgl} applies the features from both global and local patterns by using both global visual information and local region details.

\begin{table*}[tb]
\centering
\def\lw{2}
\def\ls{0.05}
\resizebox{\linewidth}{!}
{
\begin{tabular}{p{3.5cm}p{\ls cm}p{\lw cm}<{\centering}p{\lw cm}<{\centering}p{\ls cm}p{\lw cm}<{\centering}p{\lw cm}<{\centering}p{\ls cm}p{\lw cm}<{\centering}p{\lw cm}<{\centering}p{\ls cm}} 
\toprule
 \centering{Knowledge Distillation} && \multicolumn{2}{c}{NM \#5-6} && \multicolumn{2}{c}{BG \#1-2} && \multicolumn{2}{c}{CL \#1-2} \\
\cline{3-4} \cline{6-7} \cline{9-10}  \\ [-8pt]
 \centering{Function} $L_{KD}$ && GLN \cite{hou2020gln} & GaitGL \cite{lin2021gaitgl} && GLN \cite{hou2020gln} & GaitGL \cite{lin2021gaitgl} && GLN \cite{hou2020gln} & GaitGL \cite{lin2021gaitgl}   \\ 
\midrule
\multicolumn{1}{@{\hspace{1em}}l@{}}{Origin Method} && 96.5 & 97.3  && 93.1 & 94.4 && 81.5 & 83.5 \\
\multicolumn{1}{@{\hspace{2em}}l@{}}{+ RGB Body Prior} && 96.7 & 97.5 && 93.5 & 95.0 && 83.3 & 84.4 \\
\midrule
\multicolumn{1}{@{\hspace{2em}}l@{}}{+ RKD \cite{park2019relational}} &&  96.1   & 97.0   && 92.9   & 94.0   && 82.2   &  83.6  \\
\multicolumn{1}{@{\hspace{2em}}l@{}}{+ Hint \cite{romero2014fitnets}} && 96.8 & 97.4 && 93.3 & 94.4 && 83.1 & 84.0 \\
\multicolumn{1}{@{\hspace{2em}}l@{}}{+ $L_2$ \cite{ba2014deep}} && 96.7 & 96.9 && 93.2 & 94.1 && 82.9 & 84.0 \\
\multicolumn{1}{@{\hspace{2em}}l@{}}{+ NST \cite{huang2017like}} &&   96.8  &  97.2  && 93.3   &  94.4  && 82.8   &  84.1  \\
\multicolumn{1}{@{\hspace{2em}}l@{}}{+ CRD \cite{tian2019contrastive}} && 96.8 & 97.5 && 93.6 & 94.9 && 83.3 & 84.3 \\
\bottomrule
\end{tabular}
}\medskip
\caption{Ablation results for different knowledge distillation methods. Results are reported in mean accuracies on CASIA-B. `RGB body prior' indicates features used are directly encoded from the teacher model, SMPLify-X \cite{SMPL-X:2019} for selected RGB frames.} 
\label{tab:kd}
\end{table*} 

\begin{table*}[tb]
\centering
\def\lw{2}
\def\ls{0.05}
\resizebox{\linewidth}{!}
{
\begin{tabular}{p{2.5cm}p{\ls cm}p{\lw cm}<{\centering}p{\lw cm}<{\centering}p{\ls cm}p{\lw cm}<{\centering}p{\lw cm}<{\centering}p{\ls cm}p{\lw cm}<{\centering}p{\lw cm}<{\centering}p{\ls cm}} 
\toprule
 \multicolumn{1}{c}{\multirow{2}{*}{Fusion Methods}} && \multicolumn{2}{c}{NM \#5-6} && \multicolumn{2}{c}{BG \#1-2} && \multicolumn{2}{c}{CL \#1-2} \\
\cline{3-4} \cline{6-7} \cline{9-10}  \\ [-8pt]
 && GLN \cite{hou2020gln} & GaitGL \cite{lin2021gaitgl} && GLN \cite{hou2020gln} & GaitGL \cite{lin2021gaitgl} && GLN \cite{hou2020gln} & GaitGL \cite{lin2021gaitgl}   \\ 
\midrule
\multicolumn{1}{@{\hspace{1em}}l@{}}{Origin Method} && 96.5 & 97.3  && 93.1 & 94.4 && 81.5 & 83.5 \\
\multicolumn{1}{@{\hspace{2em}}l@{}}{+ MaxPool} &&  95.0  & 95.9   && 92.2   & 92.6  && 79.3   &  81.0  \\
\multicolumn{1}{@{\hspace{2em}}l@{}}{+ AvgPool} &&  96.4  & 97.2   && 93.0   & 94.4  && 82.6   &  83.7  \\
\multicolumn{1}{@{\hspace{2em}}l@{}}{+ RNN}     && 96.5 & 97.2 && 93.0 & 94.3 && 82.1 & 83.6 \\
\multicolumn{1}{@{\hspace{2em}}l@{}}{+ LSTM}    && 96.4 & 97.3 && 93.4 & 94.6 && 82.9 & 84.0 \\
\multicolumn{1}{@{\hspace{2em}}l@{}}{+ GRU}     && 96.7 & 97.5 && 93.3 & 94.6 && 83.0 & 83.9 \\
\multicolumn{1}{@{\hspace{2em}}l@{}}{+ TS}     && 96.8 & 97.7 && 93.6 & 94.8 && 83.2 & 84.1 \\
\bottomrule
\end{tabular}
}\medskip
\caption{Ablation results for different feature fusion methods for propagating inferred human body shape feature from RGB images to gait sequence on CASIA-B. TS represents temporal shifting. MaxPool and AvgPool are max pooling and average pooling respectively. Results are reported in mean accuracies.} 
\label{tab:fusion}
\end{table*}

\section{Ablation studies}

In this subsection, we discuss five different ablation studies for the composition of our model, including the choice of balancing term $\lambda_2$, the model we use for human body shape reconstruction from selected RGB images, knowledge distillation function $L_{KD}$ for knowledge transfer between two modalities, fusion method for backpropagating body shape feature from single image frames to silhouette sequence, and the ablation for feature exchange between neighbor frames.

\begin{table*}[tb]
\centering
\def\lw{2}
\def\ls{0.05}
\resizebox{\textwidth{}{}}{!}
{
\begin{tabular}{p{1.5cm}<{\centering}p{\ls cm}p{\lw cm}<{\centering}p{\lw cm}<{\centering}p{\ls cm}p{\lw cm}<{\centering}p{\lw cm}<{\centering}p{\ls cm}p{\lw cm}<{\centering}p{\lw cm}<{\centering}p{\ls cm}} 
\toprule
Balancing && \multicolumn{2}{c}{NM \#5-6} && \multicolumn{2}{c}{BG \#1-2} && \multicolumn{2}{c}{CL \#1-2} \\
\cline{3-4} \cline{6-7} \cline{9-10}  \\ [-8pt]
  Term $\lambda_2$&& GLN-HBS & GaitGL-HBS && GLN-HBS & GaitGL-HBS && GLN-HBS & GaitGL-HBS   \\ 
\midrule
0.5 && 96.6 & 97.5 && 93.4 & 94.6 && 82.8 & 84.0\\
1 && 96.8 & 97.7 && 93.6 & 94.8 && 83.2 & 84.1\\
2 && 96.6 & 97.4 && 93.5 & 94.8 && 82.6 & 83.9\\
5 && 96.2 & 97.2 && 92.9 & 94.4 && 81.6 & 83.2\\
\bottomrule
\end{tabular}
}\medskip
\caption{Ablation results for different $\lambda_2$ used for balancing $L_{KD}$ and $L_{ID}$.} 
\label{tab:lambda}
\end{table*} 

\begin{table*}[tb]
\centering
\def\lw{2}
\def\ls{0.05}
\resizebox{\textwidth}{!}
{
\begin{tabular}{p{1.25cm}p{\ls cm}p{\lw cm}<{\centering}p{\lw cm}<{\centering}p{\ls cm}p{\lw cm}<{\centering}p{\lw cm}<{\centering}p{\ls cm}p{\lw cm}<{\centering}p{\lw cm}<{\centering}p{\ls cm}} 
\toprule
 Exchange&& \multicolumn{2}{c}{NM \#5-6} && \multicolumn{2}{c}{BG \#1-2} && \multicolumn{2}{c}{CL \#1-2} \\
\cline{3-4} \cline{6-7} \cline{9-10}  \\ [-8pt]
  Ratio&& GLN-HBS & GaitGL-HBS && GLN-HBS & GaitGL-HBS && GLN-HBS & GaitGL-HBS   \\ 
\midrule
0\% &&  96.4  & 97.2   && 93.0   & 94.4  && 82.6   &  83.7  \\
10\% && 96.7 & 97.7 && 93.5 & 94.8 && 83.2 & 83.9\\
12.5\% && 96.8 & 97.7 && 93.6 & 94.8 && 83.2 & 84.1\\
25\% && 96.5 & 97.2 && 93.0 & 94.4 && 81.9 & 83.1\\
33.3\% && 95.7 & 96.8 && 92.6 & 93.5 && 81.2 & 82.9\\
\bottomrule
\end{tabular}
}\medskip
\caption{Ablations for ratio used for feature exchange in the body shape feature encoder.} 
\label{tab:ratio}
\end{table*} 

\textbf{Ablations for the Balancing Term $\lambda_2$.}
To balance the identity loss $L_{ID}$ and knowledge distillation loss $L_{KD}$, we set the balancing term follow the ablations on CASIA-B \cite{yu2006framework} for all three splits, NM, CL and BG, with GLN-HBS and GaitGL-HBS for some other variations of $\lambda_2$. We show the results in Table~\ref{tab:lambda}, where top-1 accuracy is reported excluding identical-view cases. We note that when we have the balancing term $\lambda_2$ set to 1, GLN-HBS and GaitGL-HBS both show the best performance. With $\lambda_2$ as 1, our model can find a balancing point between distinguishing different identities from silhouette sequences and transferring knowledge from inferred 3-D body shape from selected RGB frames by SMPLify-X \cite{SMPL-X:2019}.

\textbf{Body Prior Reconstruction.}
Since we need a strong human body prior to help disentangle the skinned body shape from appearance variances, to reconstruct human body prior from RGB frames, we compare the methods two skinned models, SMPL~\cite{loper2015smpl} from SMPLify~\cite{bogo2016keep} and SMPL-X~\cite{SMPL-X:2019} from SMPLify-X~\cite{SMPL-X:2019}, for 3-D body reconstruction. Compared with SMPL-X, SMPL does not require the output for human skeletons extracted by OpenPose~\cite{cao2019openpose}. We assess both methods on the CASIA-B dataset for three settings with GLN. For SMPLify, the average accuracies are 96.7, 93.4 and 82.6 for NM, BG and CL, respectively, while for SMPLify-X, the average accuracies are 96.7, 93.6 and 83.2. 
Although SMPL shows some improvement compared with GLN without 3-D human body shape, the inaccurate reconstructions from SMPLify make the network unable to distinguish between body shapes and appearance variances, making it unable to beat SMPLify-X reconstructions.

\textbf{Knowledge Distillation.} We show the results for different knowledge distillation methods \cite{park2019relational,ba2014deep,romero2014fitnets,tian2019contrastive,huang2017like}, in addition to the experiment directly using the feature output from the teacher network, in Table~\ref{tab:kd}. Since GLN and GaitGL are the two state-of-the-art methods with the best performance in Table~\ref{tab:casiab-1}, we compare several knowledge distillation methods on all three variations of the CASIA-B dataset for GLN and GaitGL with SMPLify-X as the 3-D human body shape reconstruction model for RGB images. Among all the knowledge distillation methods, CRD shows the best performance, and we choose to use CRD as our $L_{KD}$ for features of 3-D body shape transfer from RGB frame $s_r$ to gait sequence $g$. In addition, we also note from the table that using the distilled feature from CRD is comparable to the body prior directly extracted from selected RGB frames by the teacher network, SMPLify-X \cite{SMPL-X:2019}, and even better at some splits. With knowledge distillation, body shape from gait sequence can be more stable than using a single RGB image for reconstruction.

\textbf{Fusion.} In addition to the method selection for knowledge distillation, we further show different methods for propagating the single frame RGB features to gait sequences in Table~\ref{tab:fusion}. We assess different fusion methods on CASIA-B using CRD for knowledge distillation and transfer. In addition to the temporal shifting, annotated as TS in the table, we assess two pooling and three RNN variations. We note that the max-pooling results are worse than the original methods, indicating that the model starts overfitting on a few frames. Compared with average pooling and three RNN variations, temporal shifting introduces the most significant improvement. The ability to propagate single frame information back to all frames and exchange the features between nearby frames introduce more stability and consistency for knowledge transfer.

\textbf{Ablation for the Ratio of feature exchange.} 
To temporally shift the features extracted from the body shape feature encoder in the gait feature extraction branch, we follow \cite{lin2019tsm} to set the ratio of feature exchange to 12.5\%. This number indicates that we use 75\% of features from the current frame, 12.5\% from future frames, and 12.5\% from the previous frame for the next step's convolution operation. We further research several different ratios of feature exchange in Table~\ref{tab:ratio}. We note that when we exchange 12.5\%, following \cite{lin2019tsm}, as what we did in the main paper, our models show the best performance. When we increase the exchange ratio to 33.3\%, the feature from the current frame is the same amount as the feature from the previous and next frames. At this ratio, the model cannot extract enough information from the current frame to identify the person in the sequence. When we set the exchange ratio as 0\%, the model degenerates to the average pooling case, where no features are exchanged for temporal fusion before the average pooling layer.

\section{Visualizations for Inferred Body Shapes.}
We visualize some reconstructions of human body shapes to assess the quality of inferred body shape $v_{bs}$ from silhouette sequences. We convert $v_{bs}$ to the form of the body shape feature $\beta$ used by the skinned human body reconstruction model SMPL-X \cite{SMPL-X:2019} in the reverse way that we normalize it. Since we do not predict human poses $\theta$ from silhouette with our model, we plot body shapes as T-poses for all reconstructions. We choose two examples in the test set of CASIA-B \cite{yu2006framework} with all three variants. To assess the stability among different camera positions, we select four camera positions for each subject: 0\degree, 36\degree, 72\degree\ and 108\degree.

\begin{figure*}
\centering
\resizebox{\linewidth}{!}
{
\centering
\def\lw{2}
\def\imw{2.2}
\def\ls{0.05}
\begin{tabular}{p{\imw cm}<{\centering}p{\ls cm}p{\imw cm}<{\centering}p{\imw cm}<{\centering}p{\ls cm}p{\imw cm}<{\centering}p{\imw cm}<{\centering}p{\ls cm}p{\imw cm}<{\centering}p{\imw cm}<{\centering}}
\toprule
        Camera && \multicolumn{2}{c}{NM} && \multicolumn{2}{c}{CL} && \multicolumn{2}{c}{BG}\\
        \cline{3-4} \cline{6-7} \cline{9-10}  \\ [-8pt]
        Viewpoints &&  RGB &  silhouette&&  RGB &  silhouette&&  RGB &  silhouette\\
        \midrule
        
        \includegraphics[width=\lw cm]{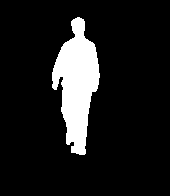} && 
        \includegraphics[width=\lw cm]{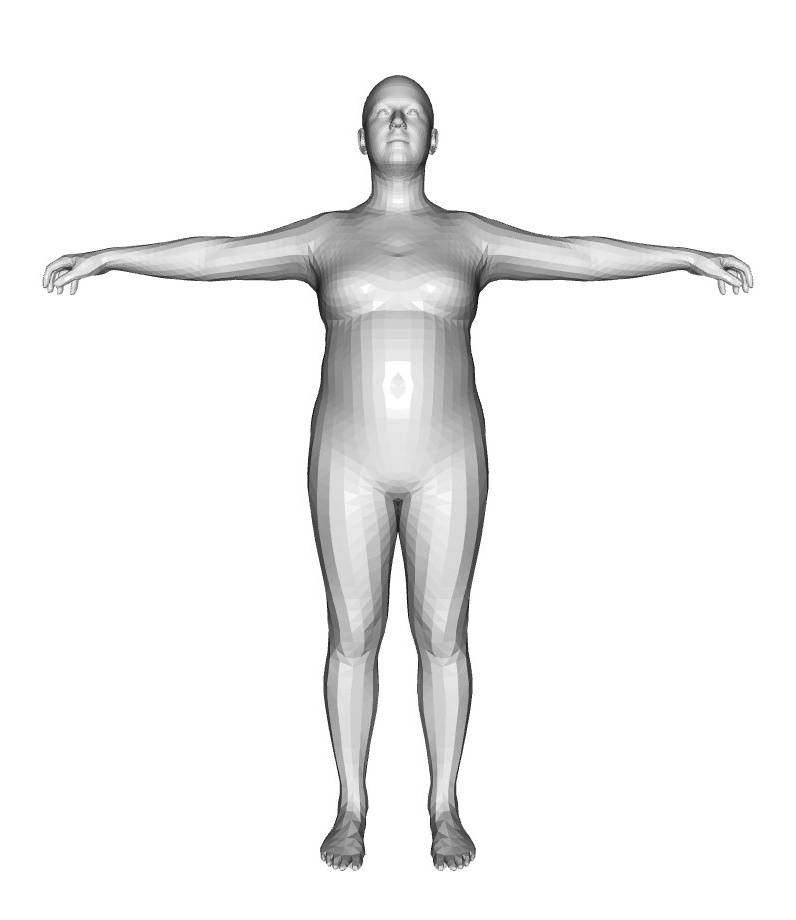} & 
        \includegraphics[width=\lw cm]{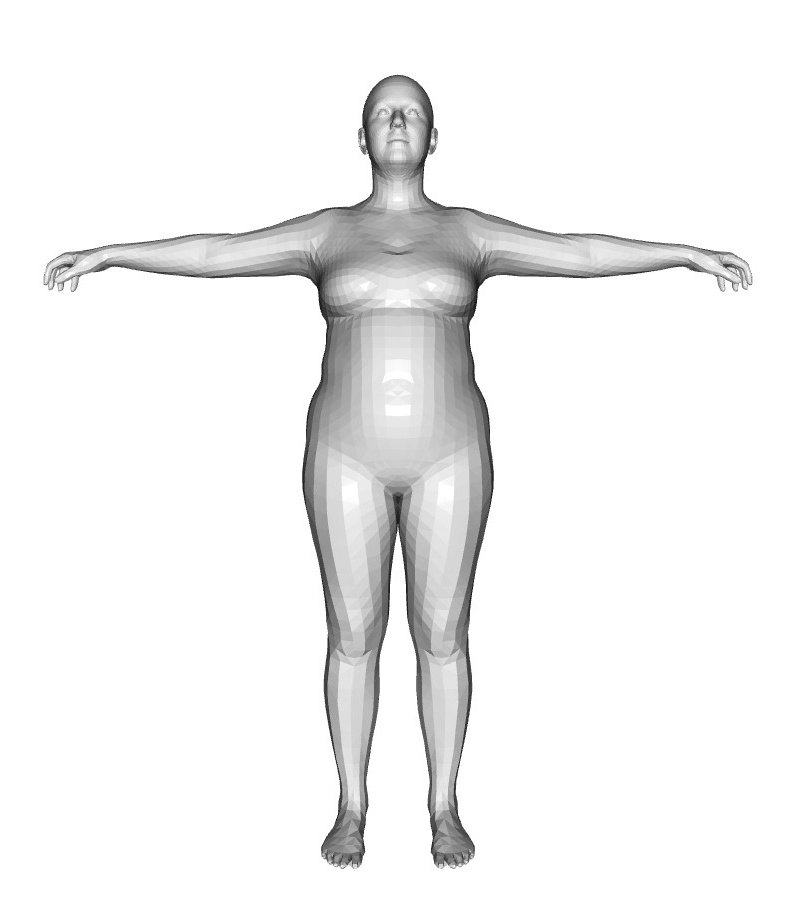} &&
        \includegraphics[width=\lw cm]{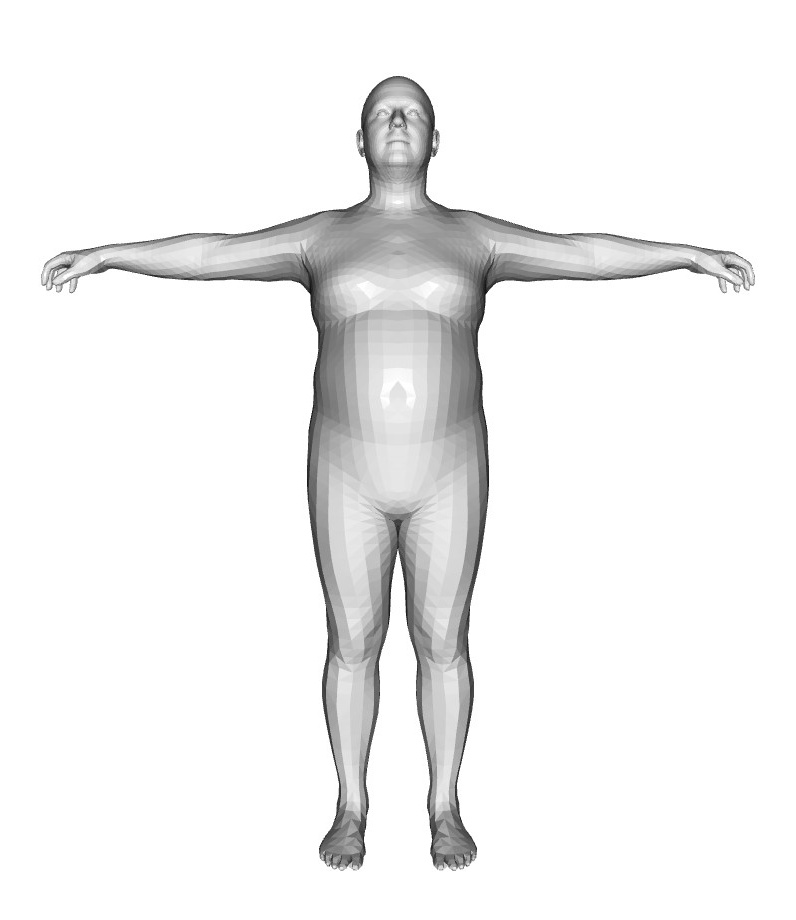} & 
        \includegraphics[width=\lw cm]{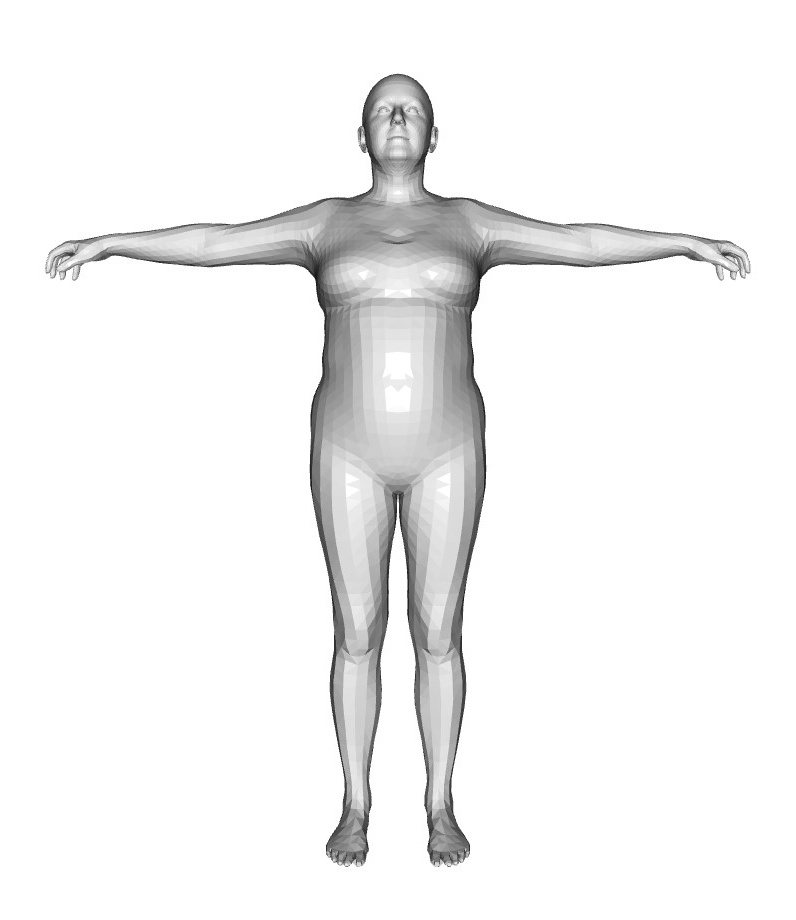} &&
        \includegraphics[width=\lw cm]{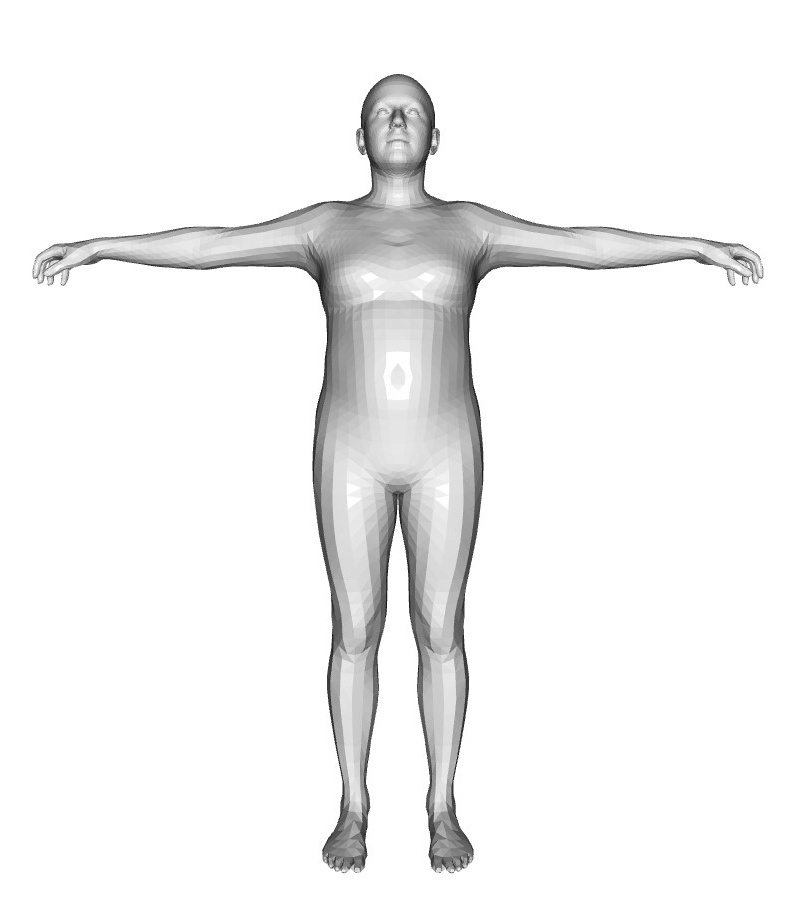} & 
        \includegraphics[width=\lw cm]{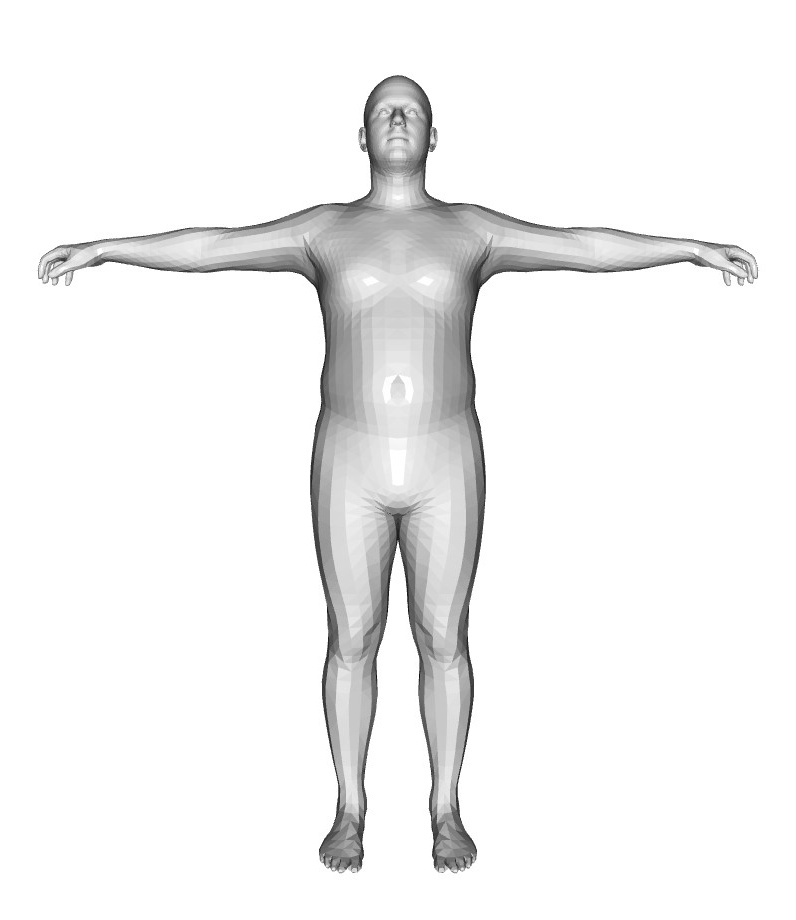}\\

        \includegraphics[width=\lw cm]{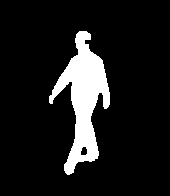} && 
        \includegraphics[width=\lw cm]{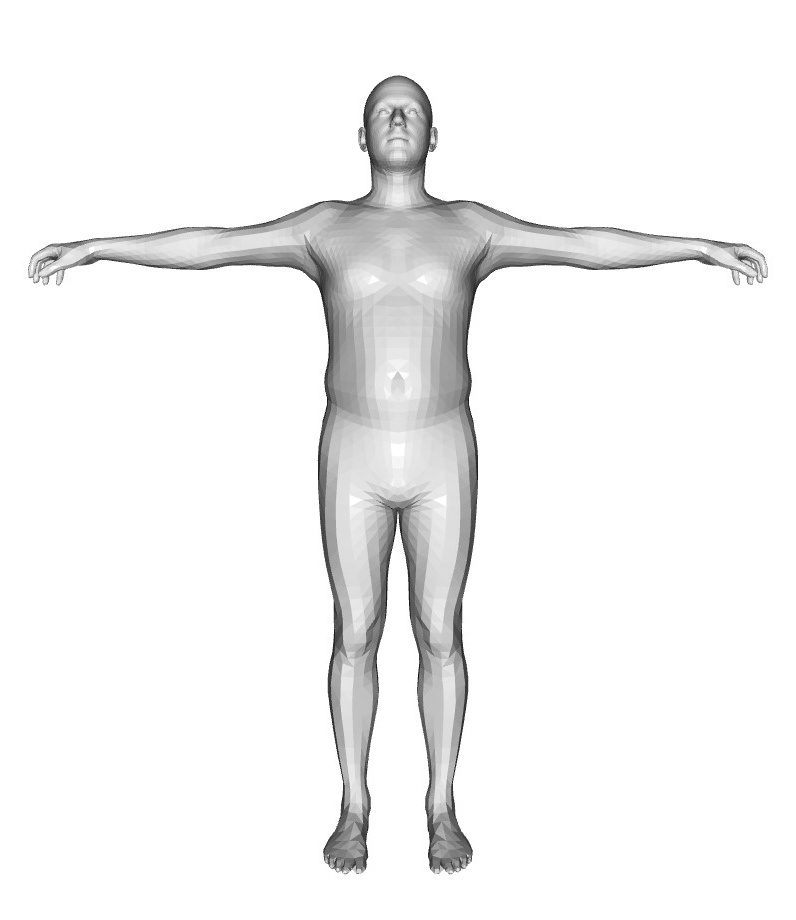} & 
        \includegraphics[width=\lw cm]{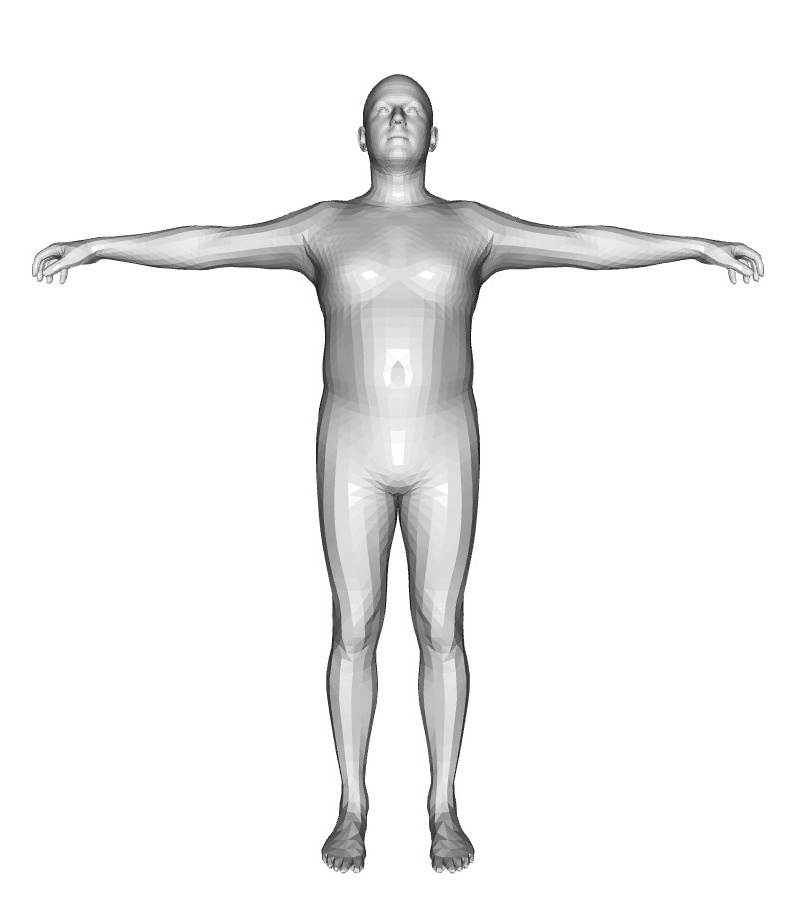} &&
        \includegraphics[width=\lw cm]{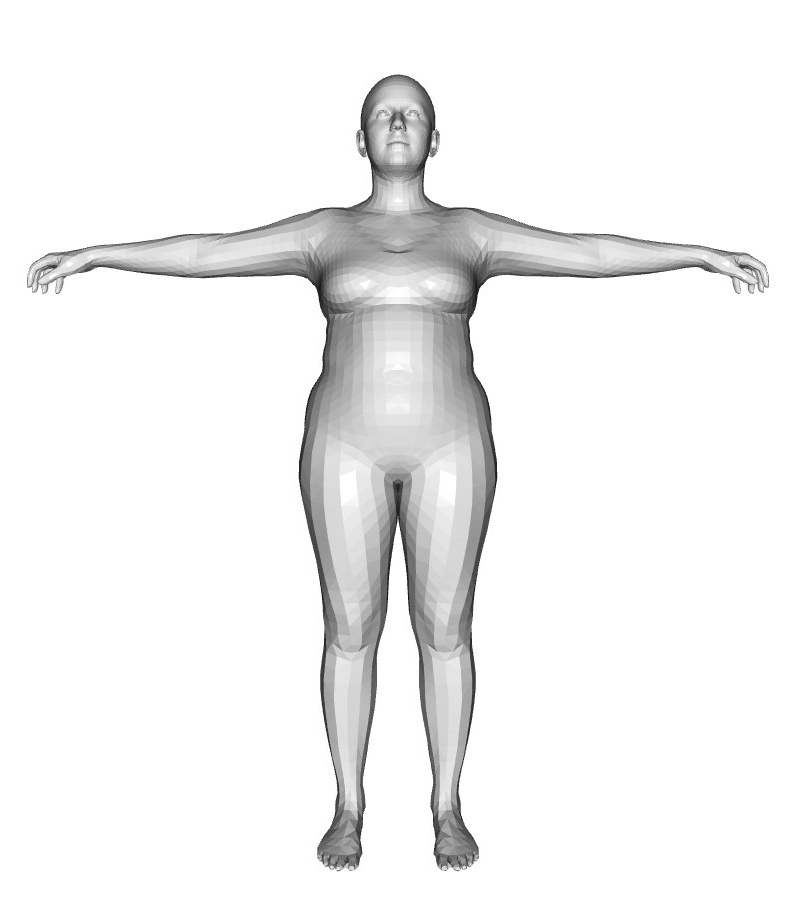} & 
        \includegraphics[width=\lw cm]{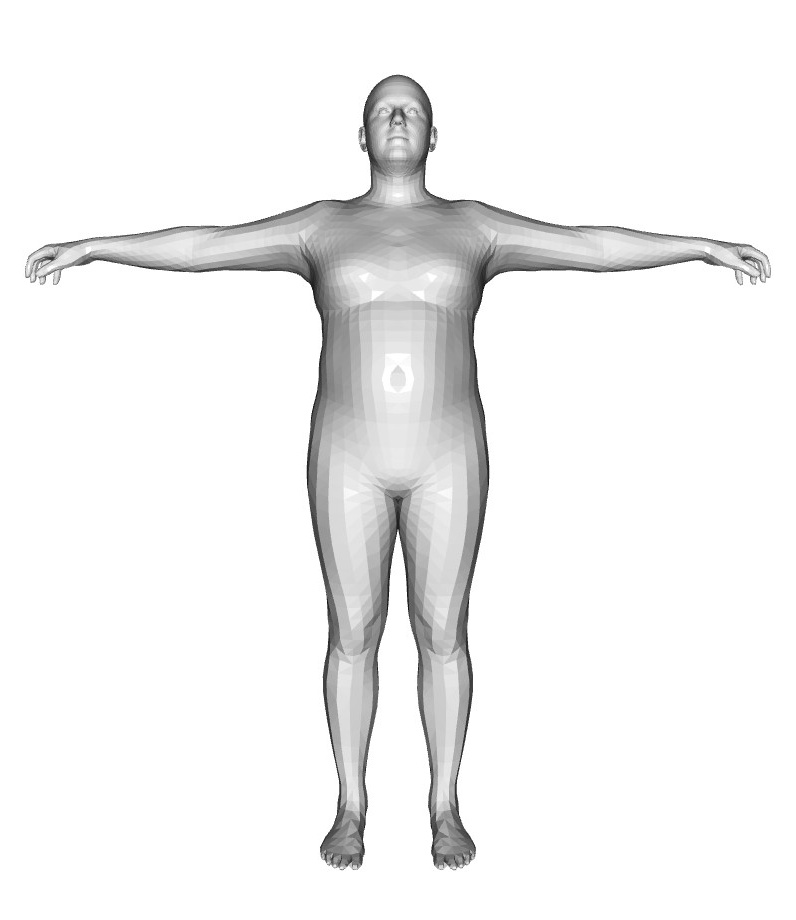} &&
        \includegraphics[width=\lw cm]{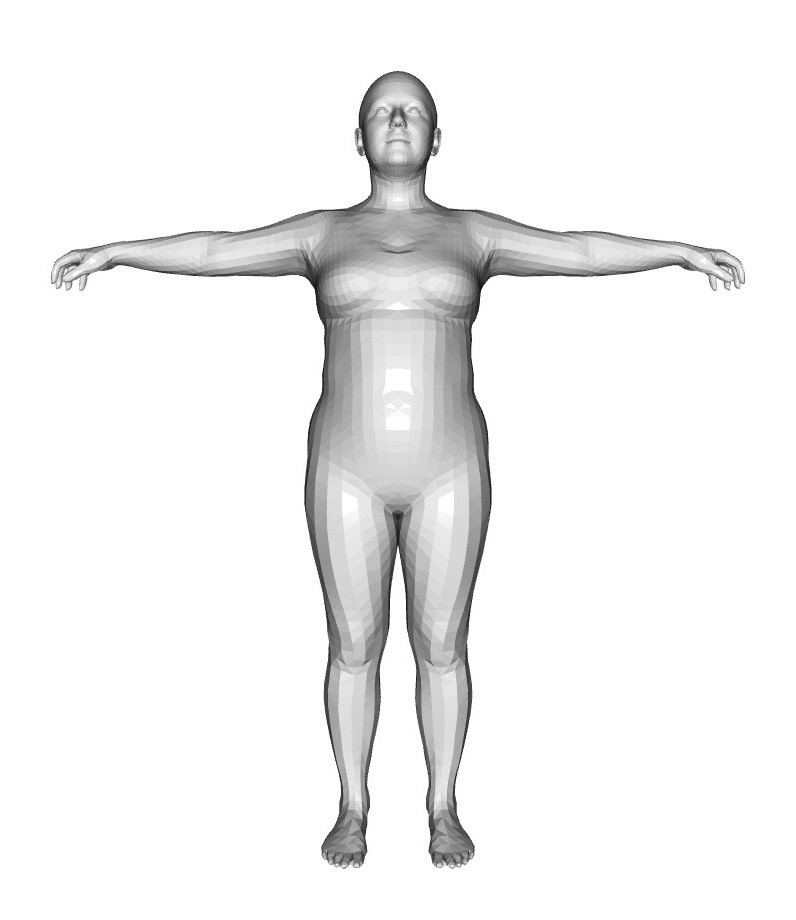} & 
        \includegraphics[width=\lw cm]{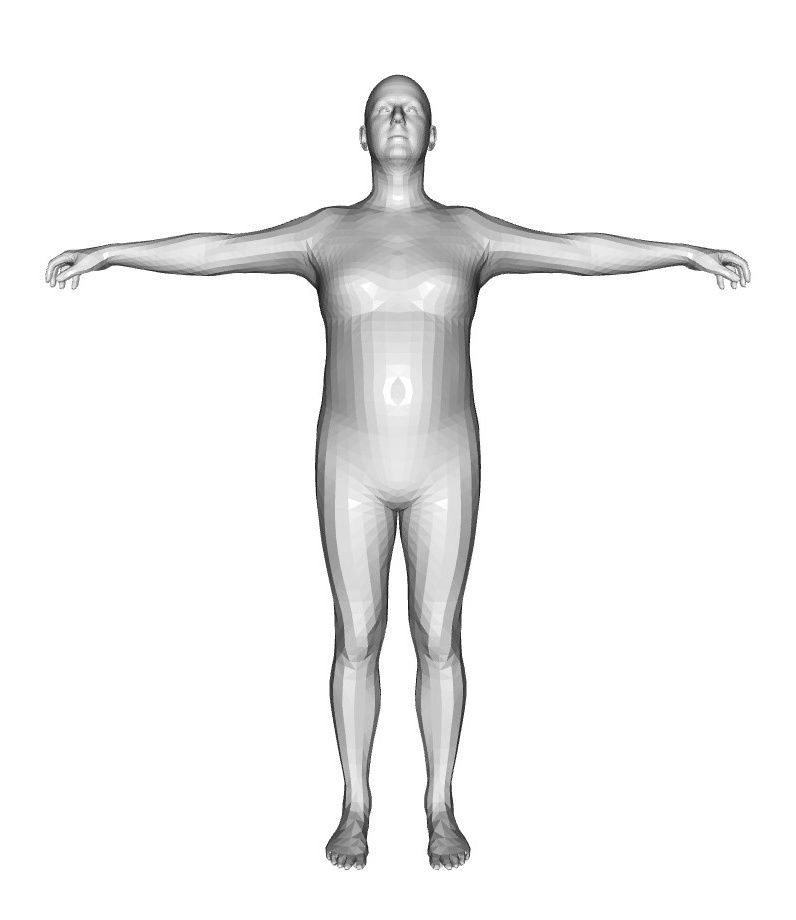}\\

        \includegraphics[width=\lw cm]{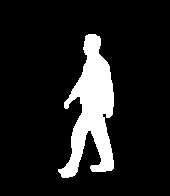} && 
        \includegraphics[width=\lw cm]{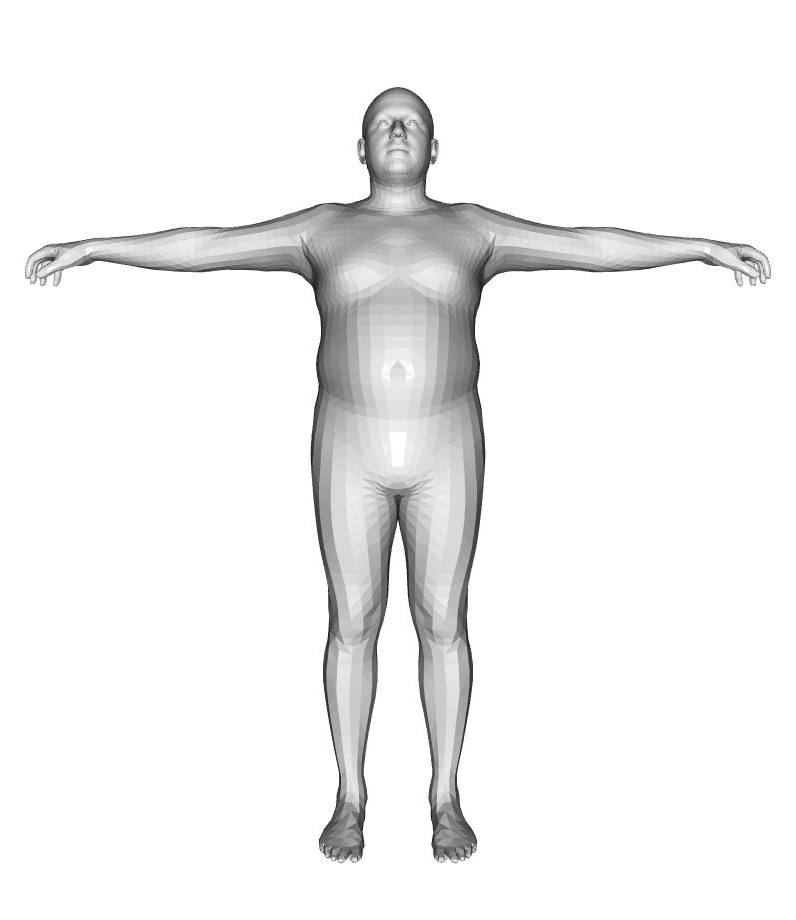} & 
        \includegraphics[width=\lw cm]{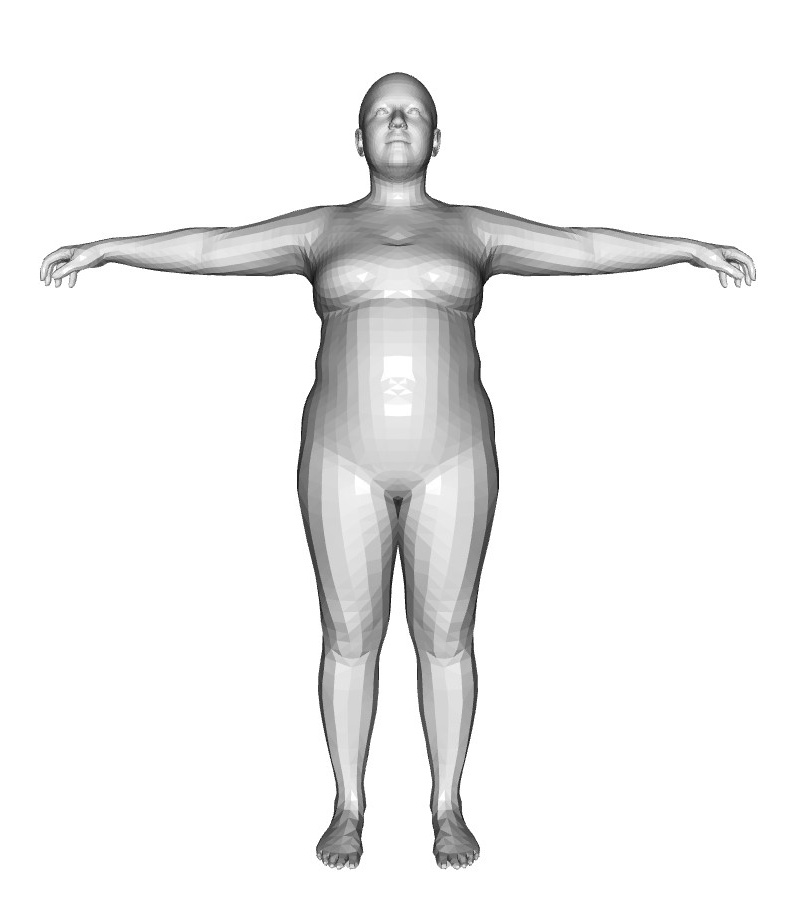} &&
        \includegraphics[width=\lw cm]{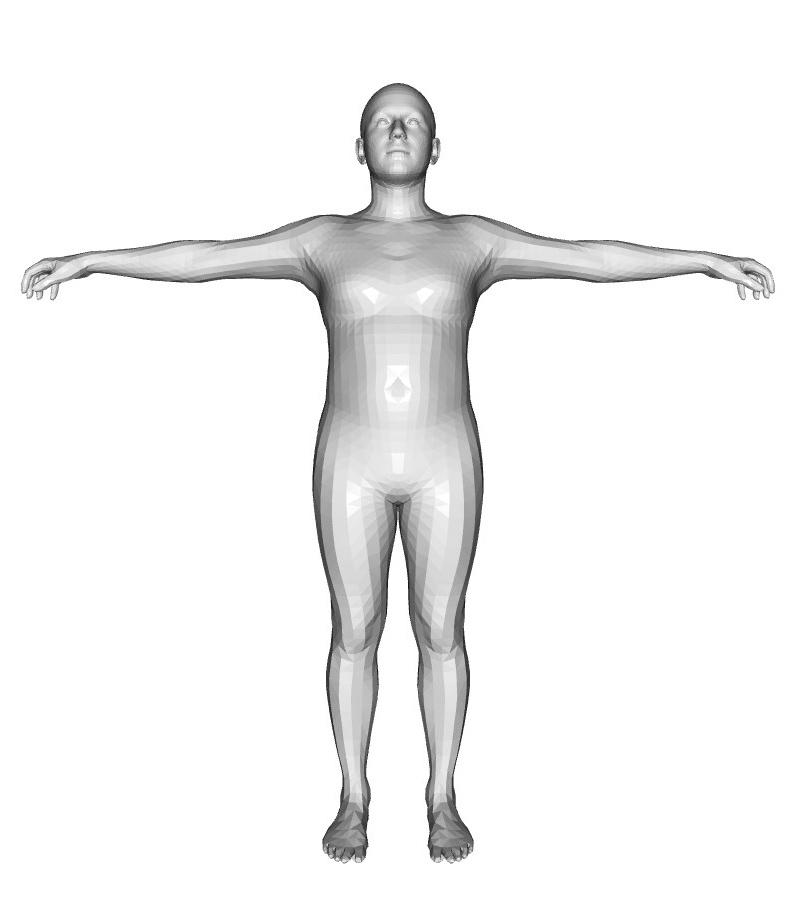} & 
        \includegraphics[width=\lw cm]{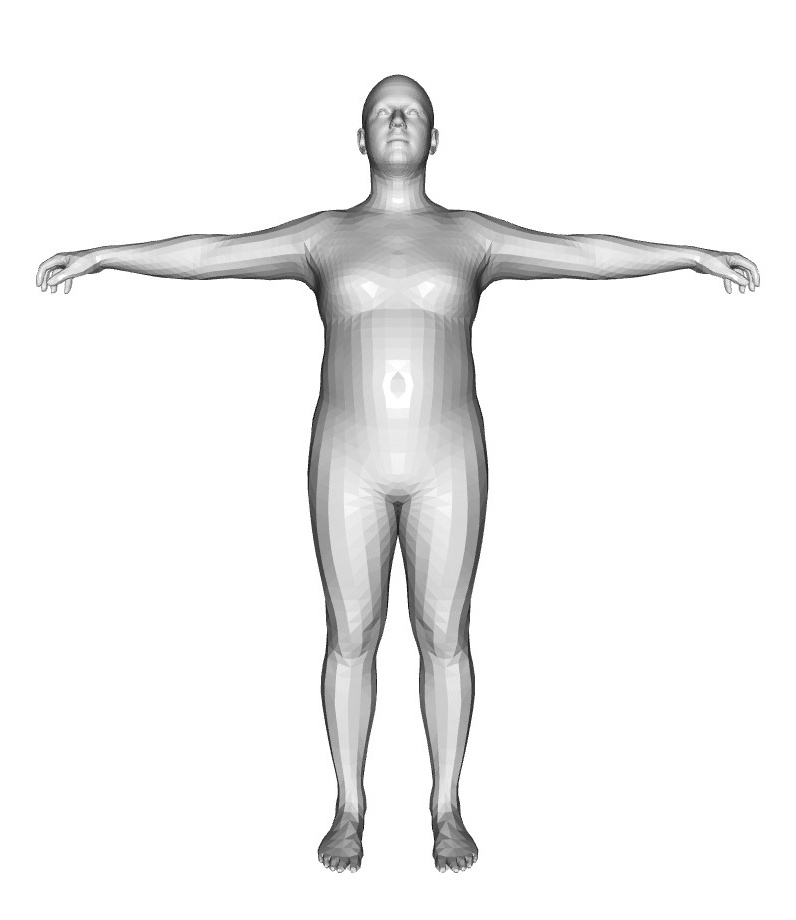} &&
        \includegraphics[width=\lw cm]{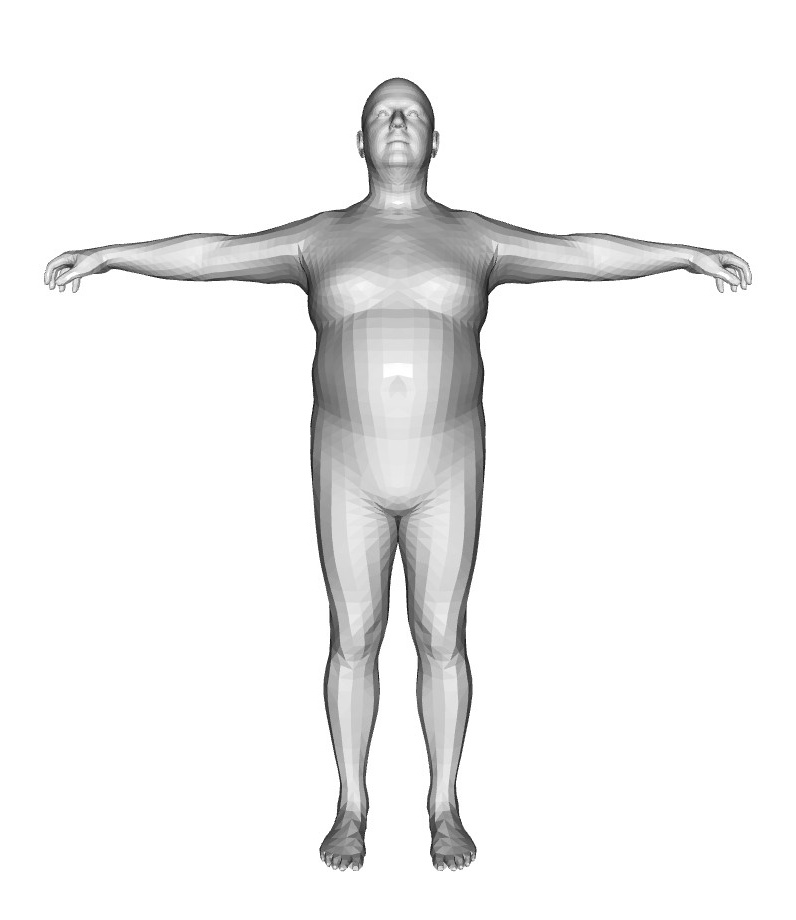} & 
        \includegraphics[width=\lw cm]{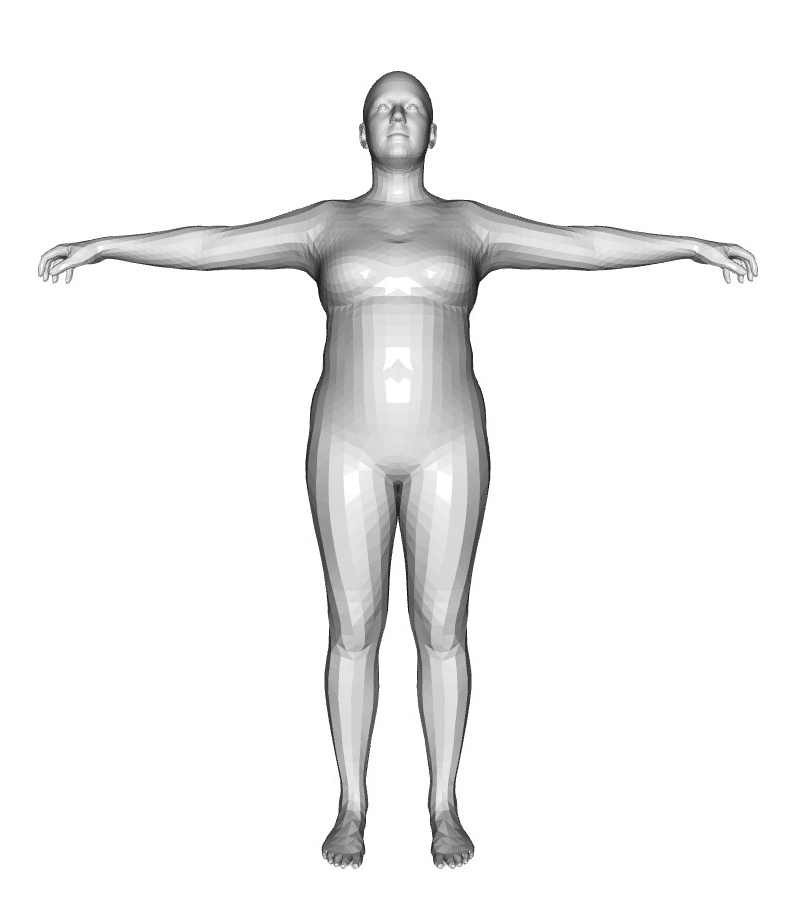}\\

        \includegraphics[width=\lw cm]{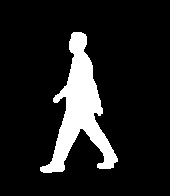} &&
        \includegraphics[width=\lw cm]{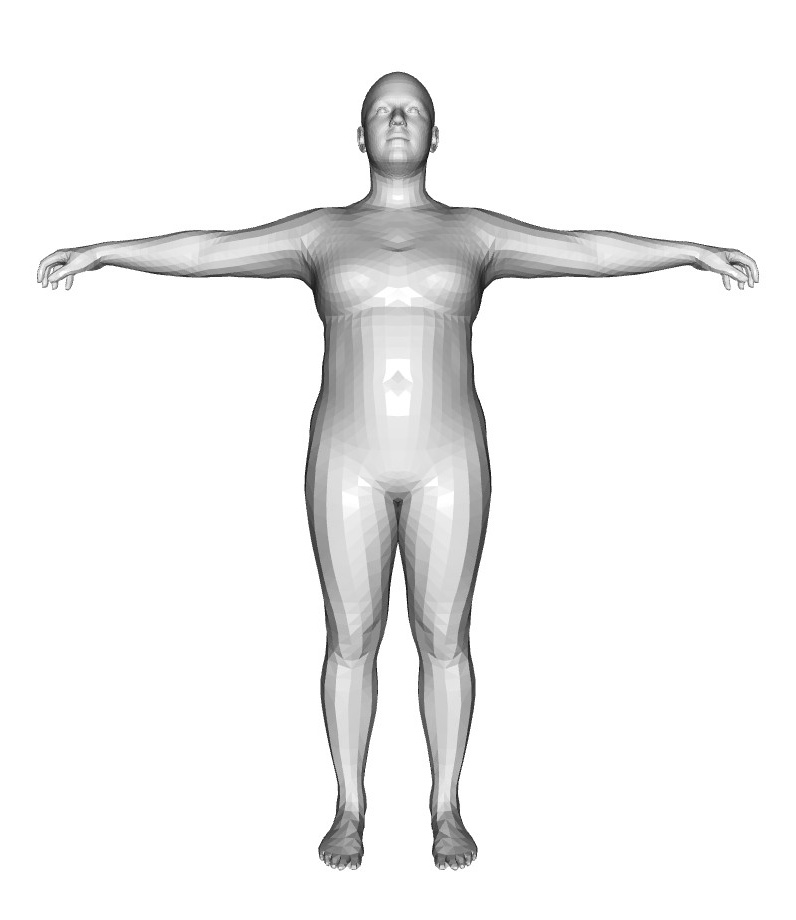} & 
        \includegraphics[width=\lw cm]{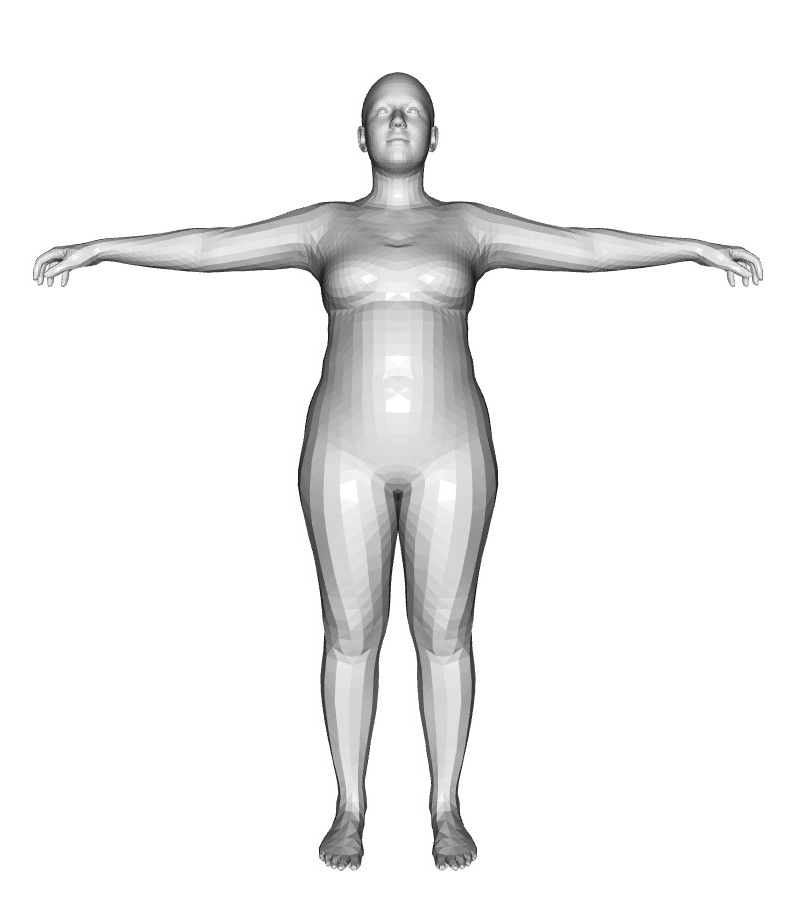} &&
        \includegraphics[width=\lw cm]{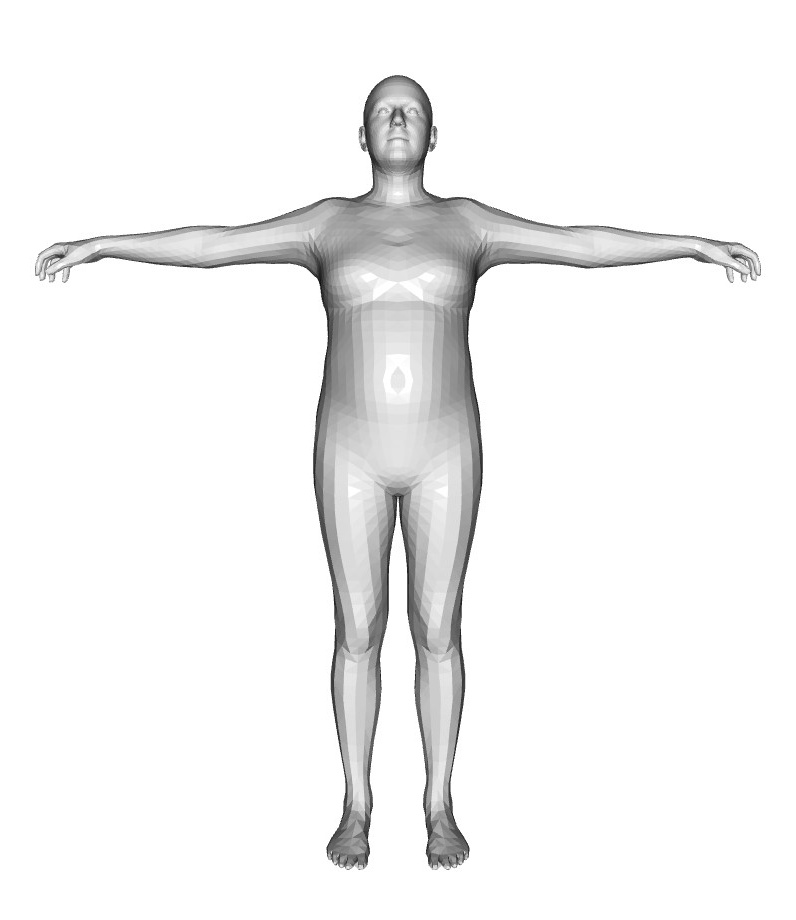} & 
        \includegraphics[width=\lw cm]{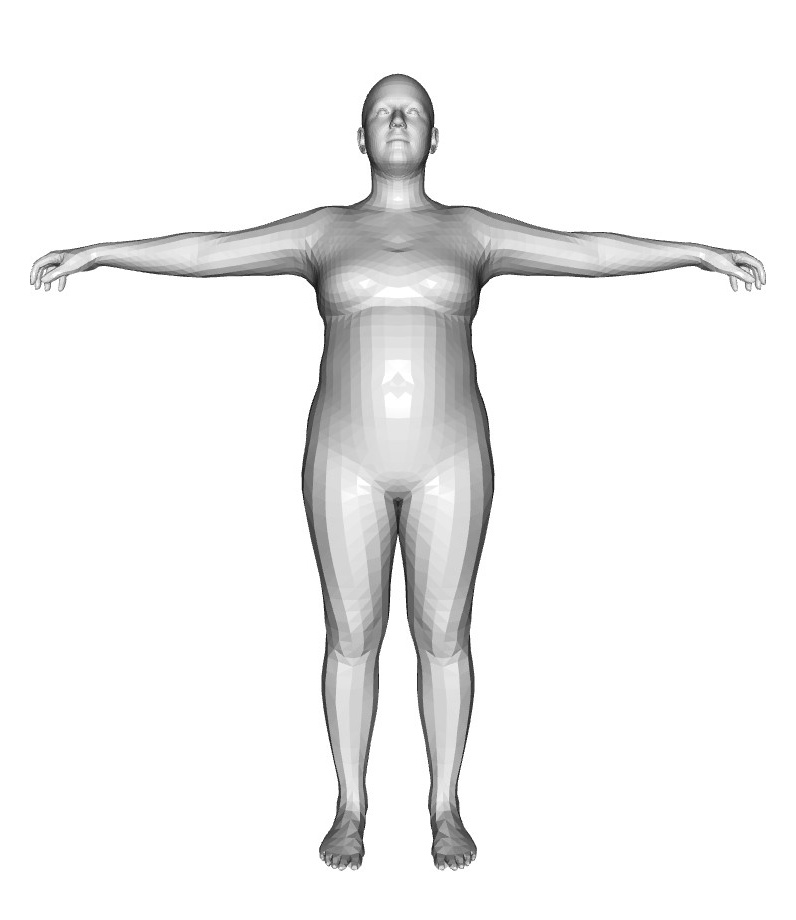} &&
        \includegraphics[width=\lw cm]{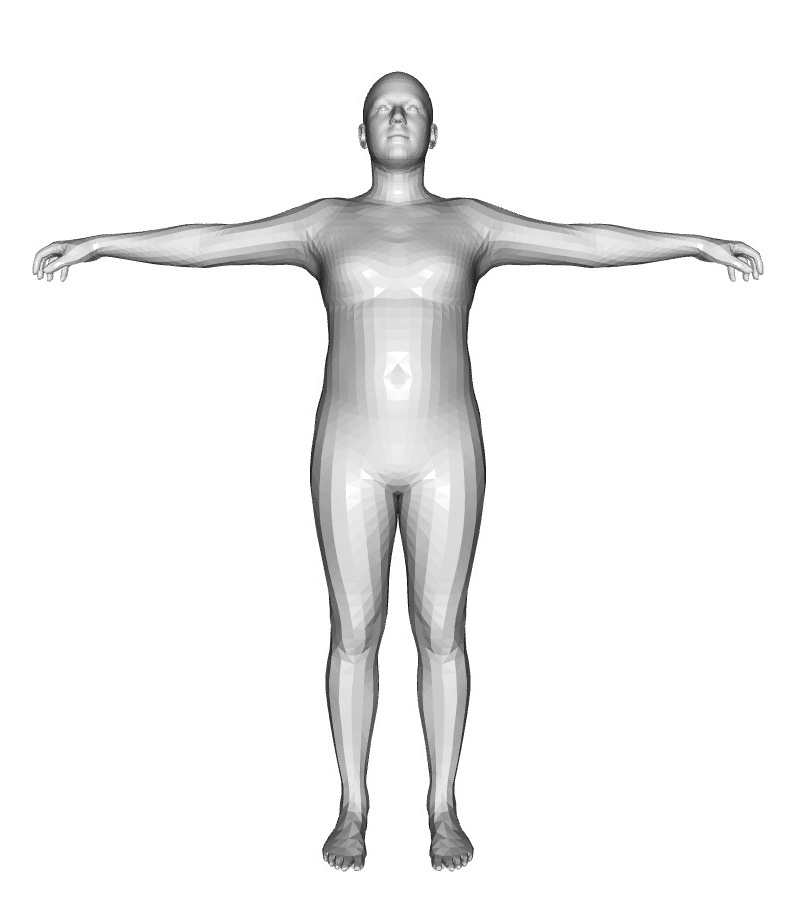} & 
        \includegraphics[width=\lw cm]{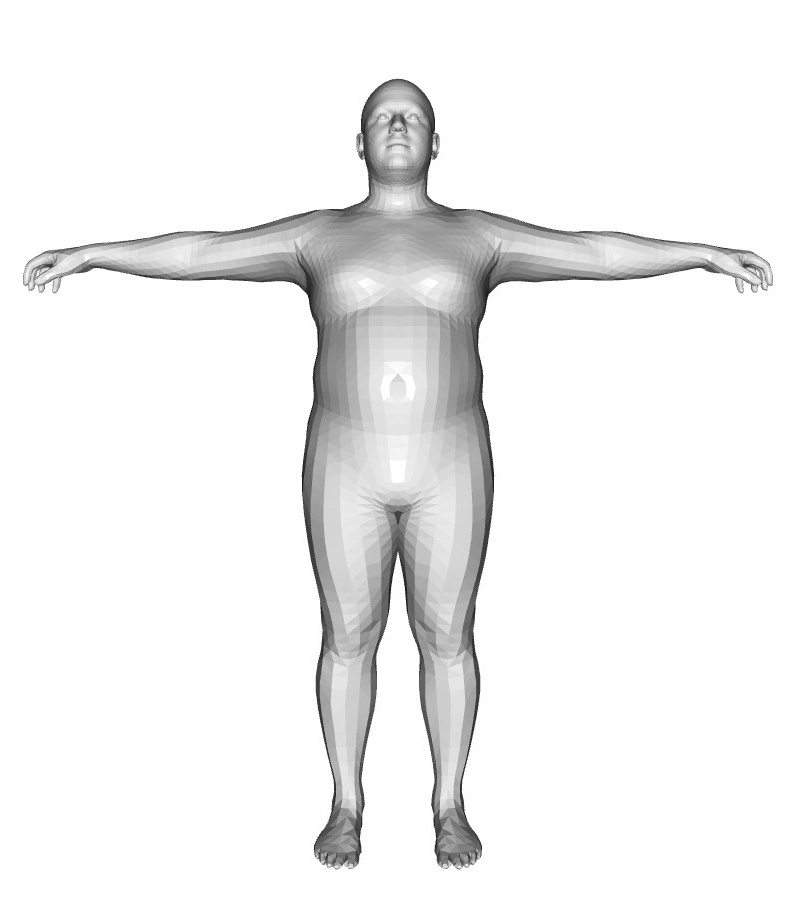}\\
        
        \midrule
        
        \includegraphics[width=\lw cm]{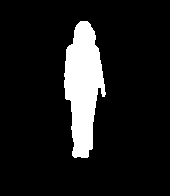} && 
        \includegraphics[width=\lw cm]{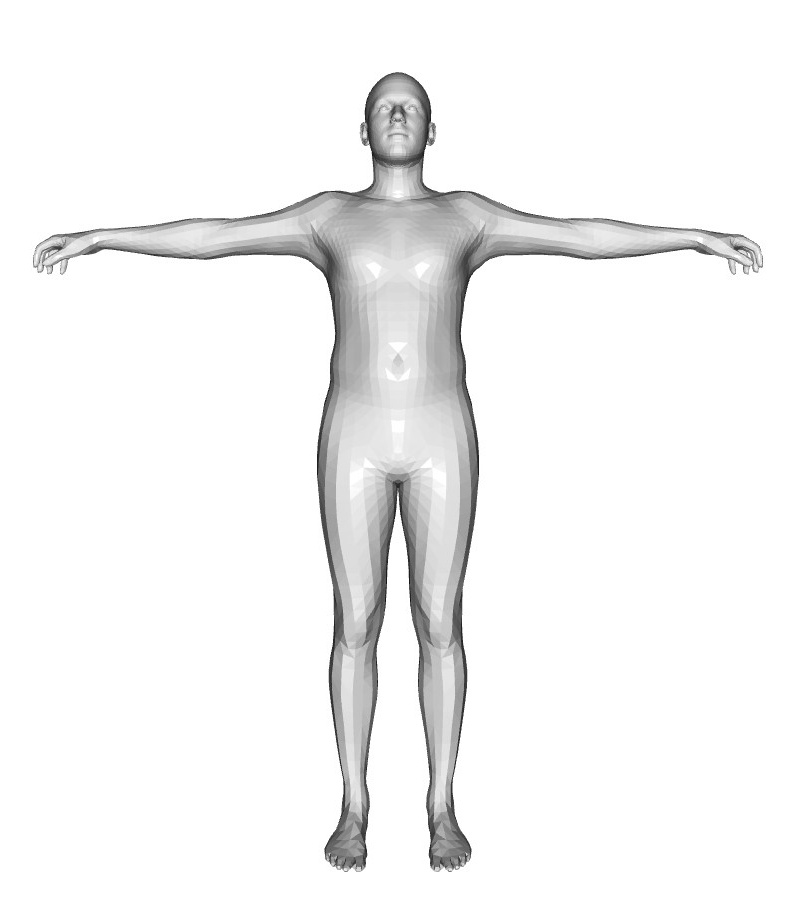} & 
        \includegraphics[width=\lw cm]{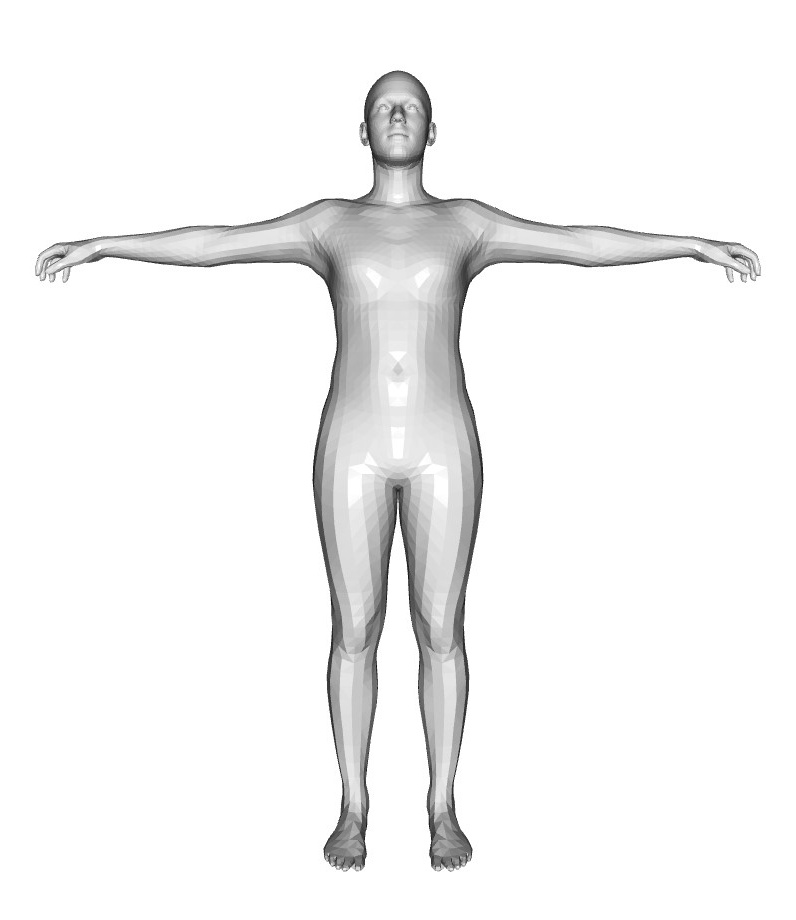} &&
        \includegraphics[width=\lw cm]{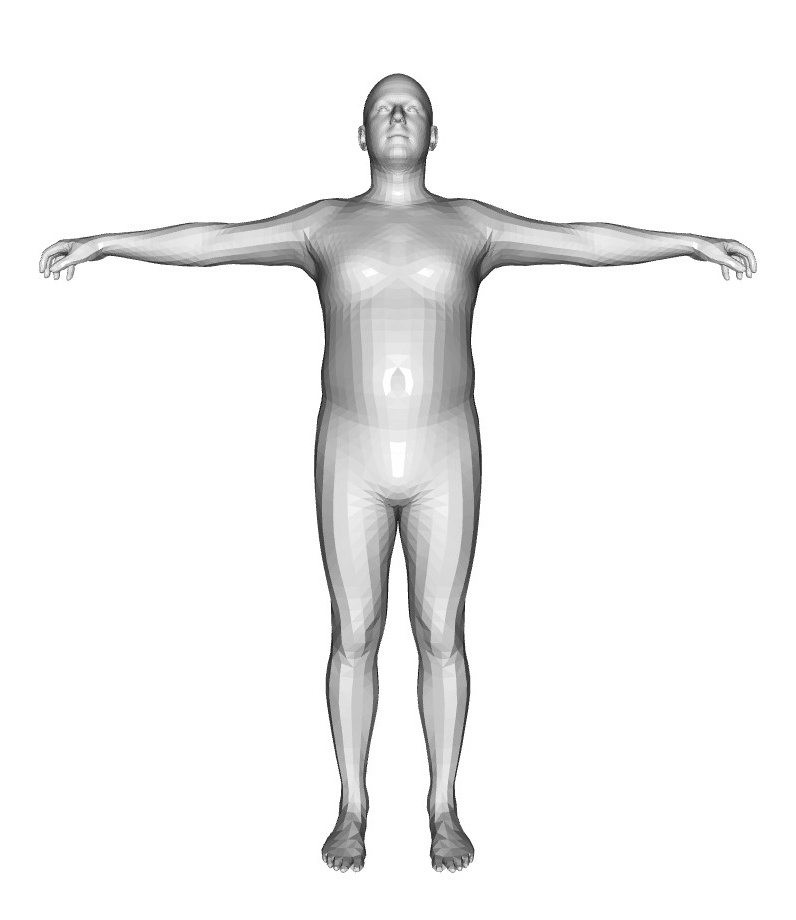} & 
        \includegraphics[width=\lw cm]{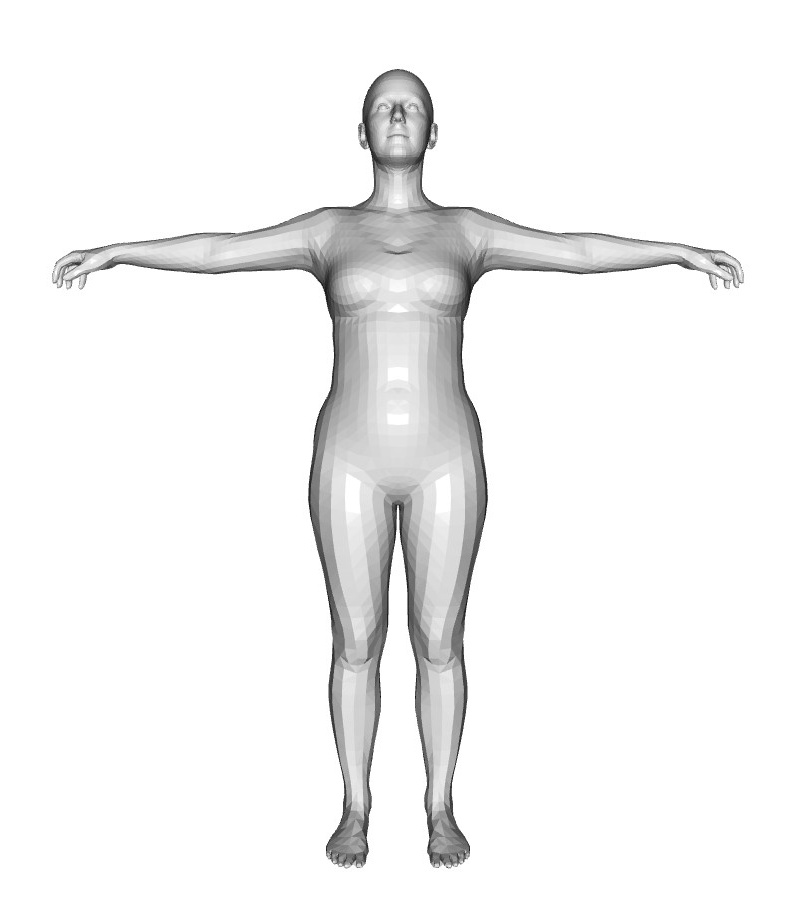} &&
        \includegraphics[width=\lw cm]{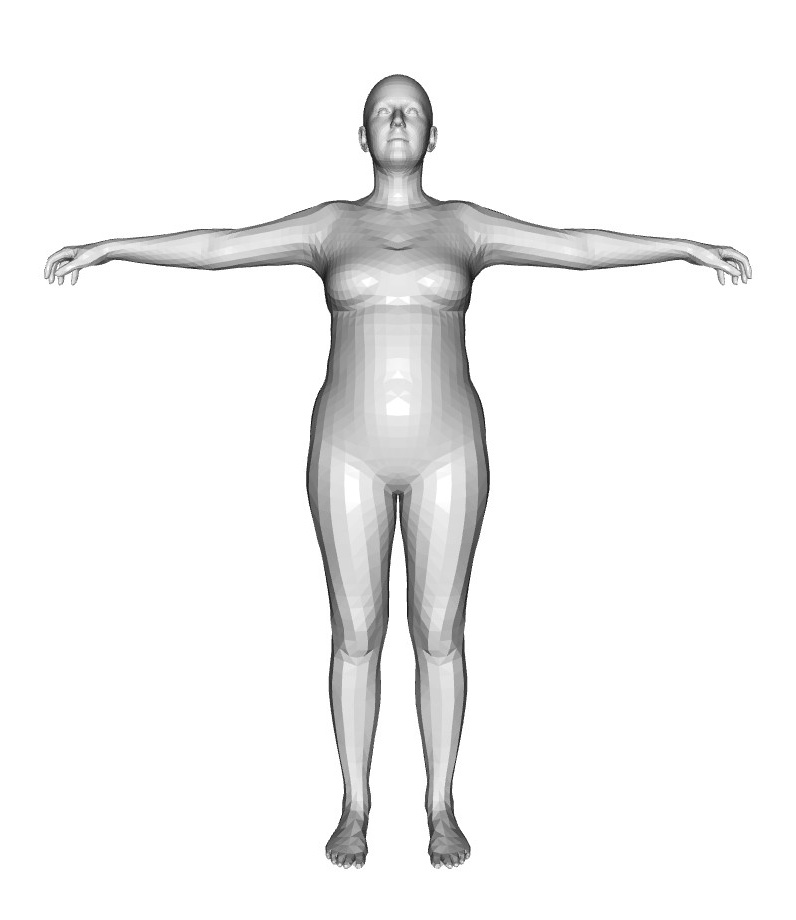} & 
        \includegraphics[width=\lw cm]{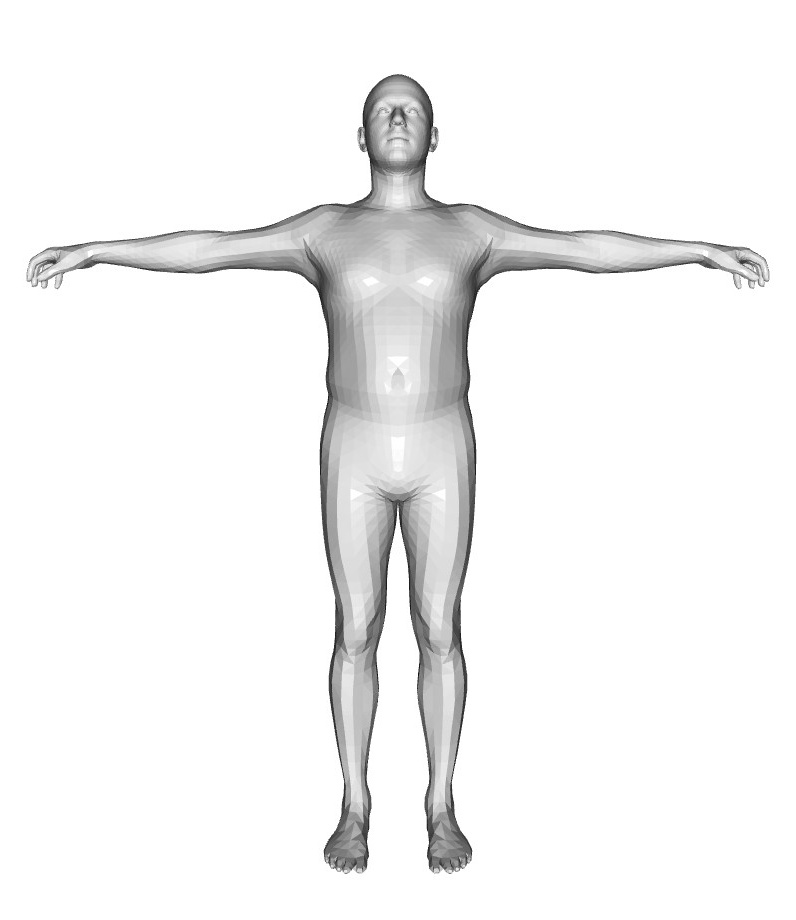}\\

        \includegraphics[width=\lw cm]{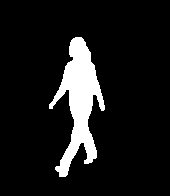} && 
        \includegraphics[width=\lw cm]{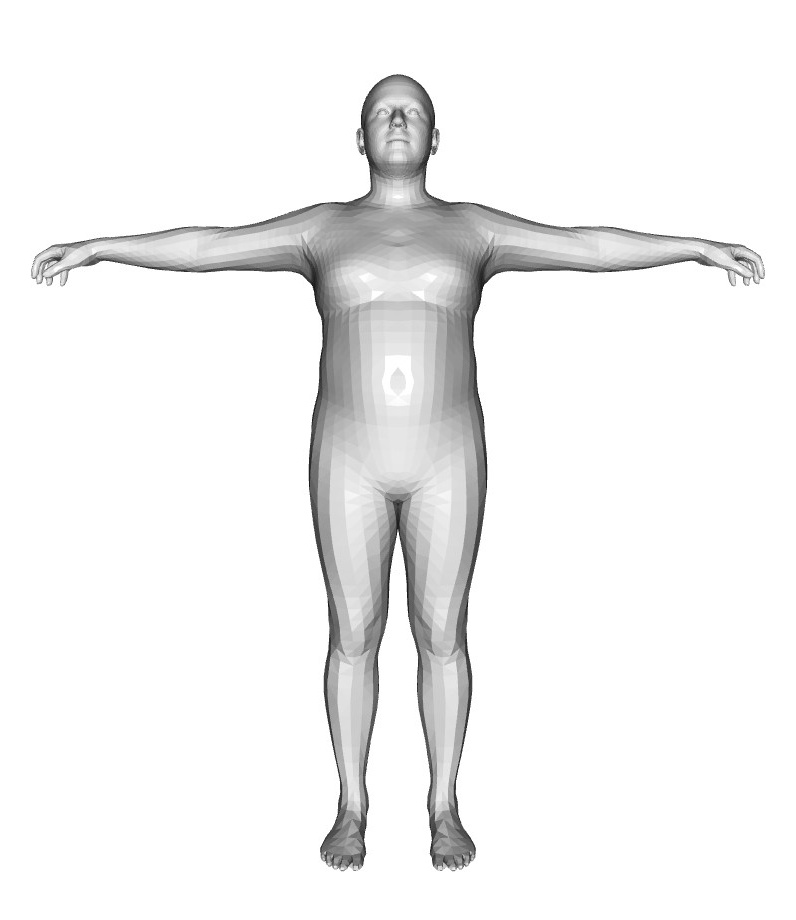} & 
        \includegraphics[width=\lw cm]{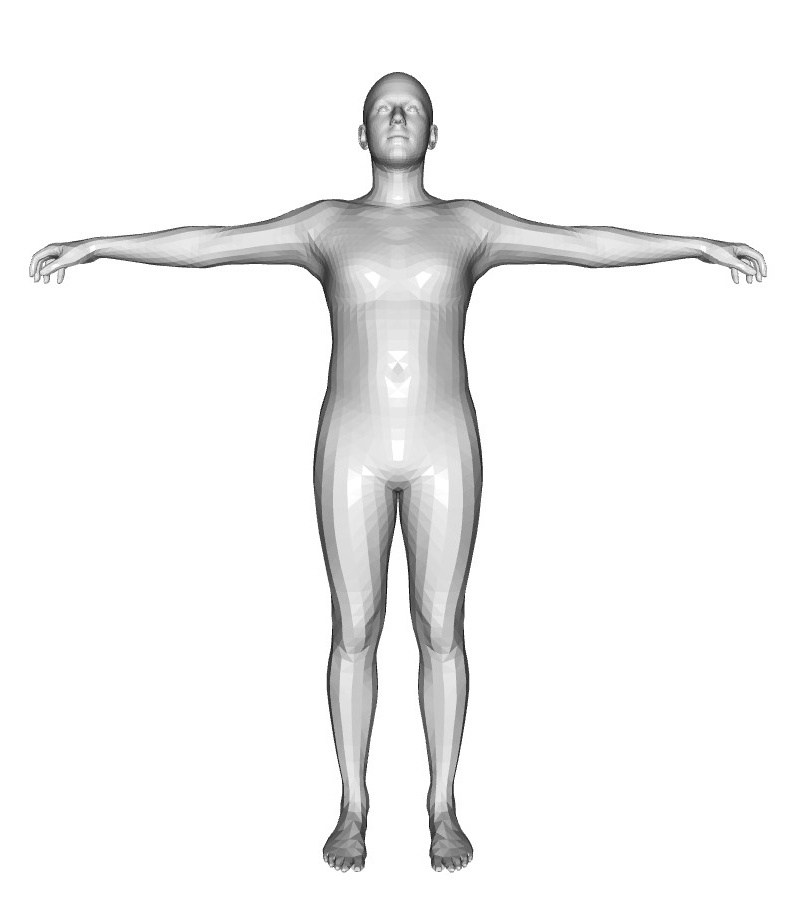} &&
        \includegraphics[width=\lw cm]{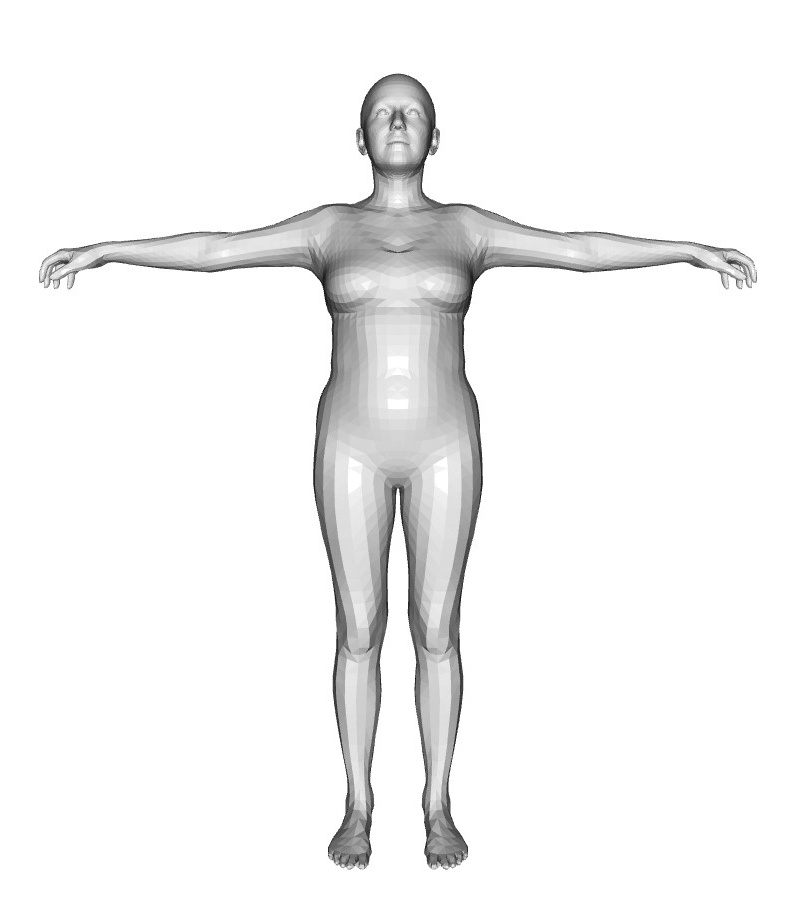} & 
        \includegraphics[width=\lw cm]{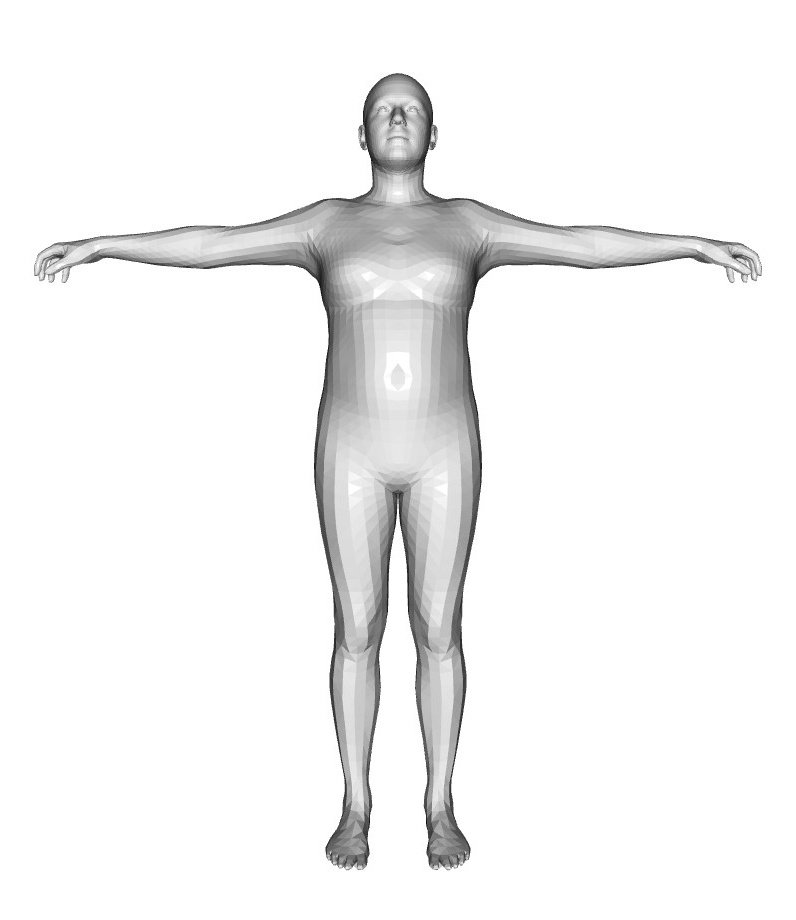} &&
        \includegraphics[width=\lw cm]{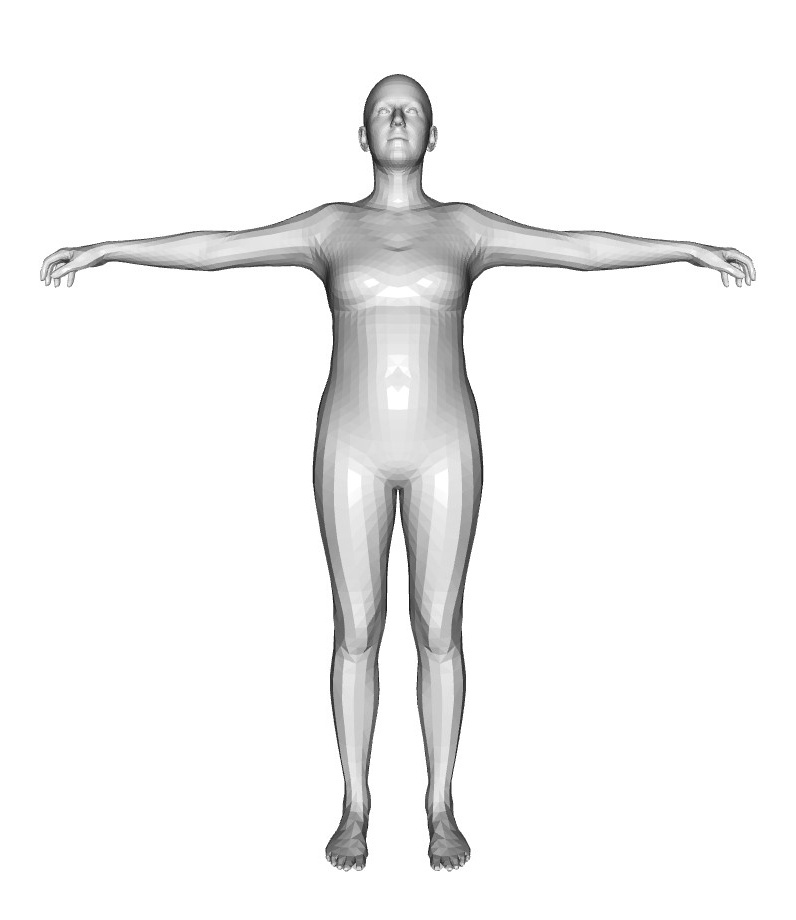} & 
        \includegraphics[width=\lw cm]{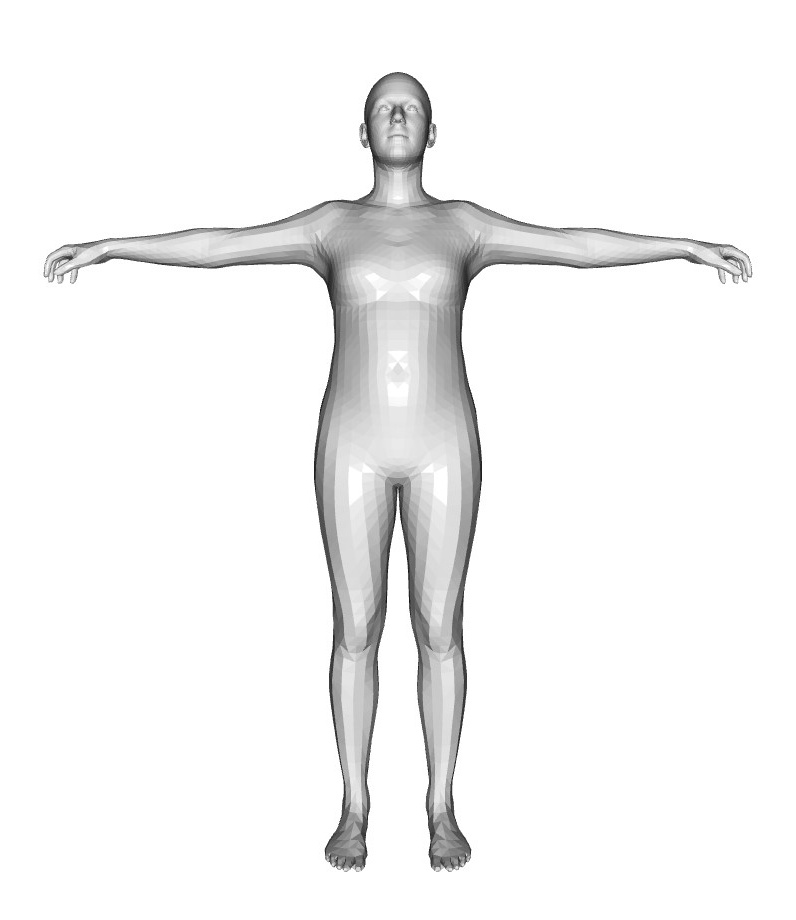}\\

        \includegraphics[width=\lw cm]{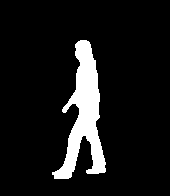} && 
        \includegraphics[width=\lw cm]{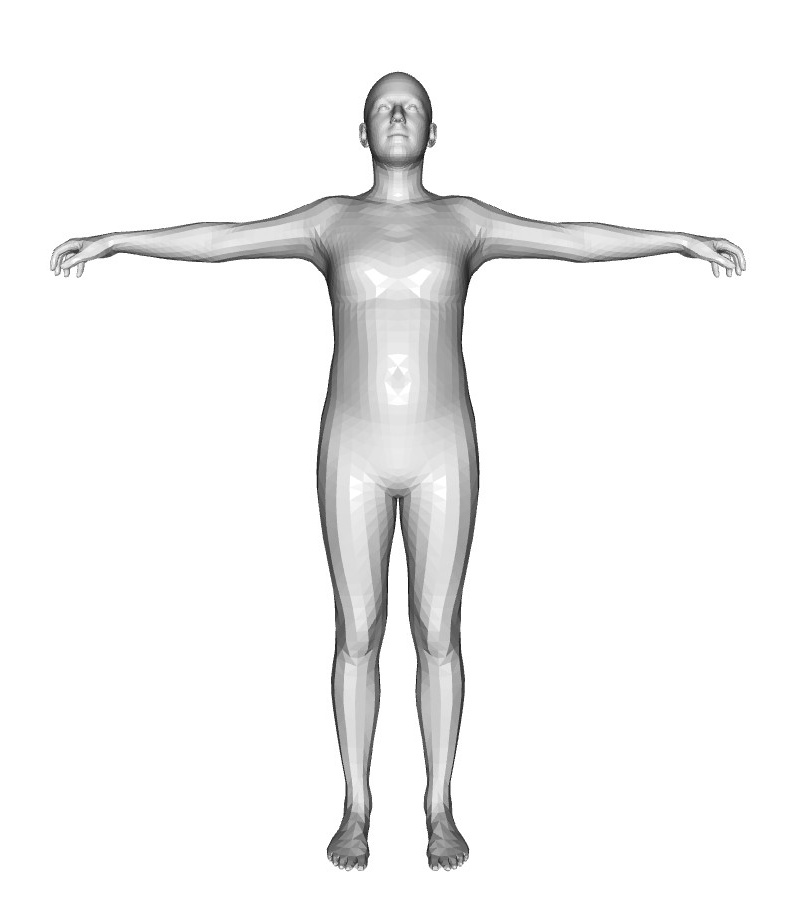} & 
        \includegraphics[width=\lw cm]{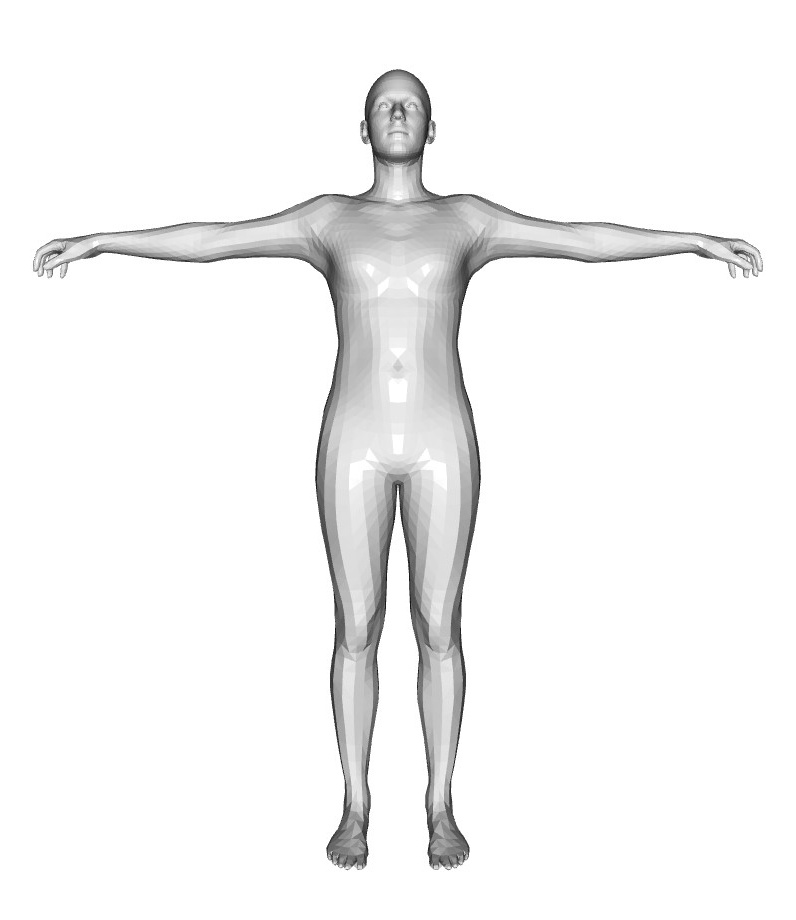} &&
        \includegraphics[width=\lw cm]{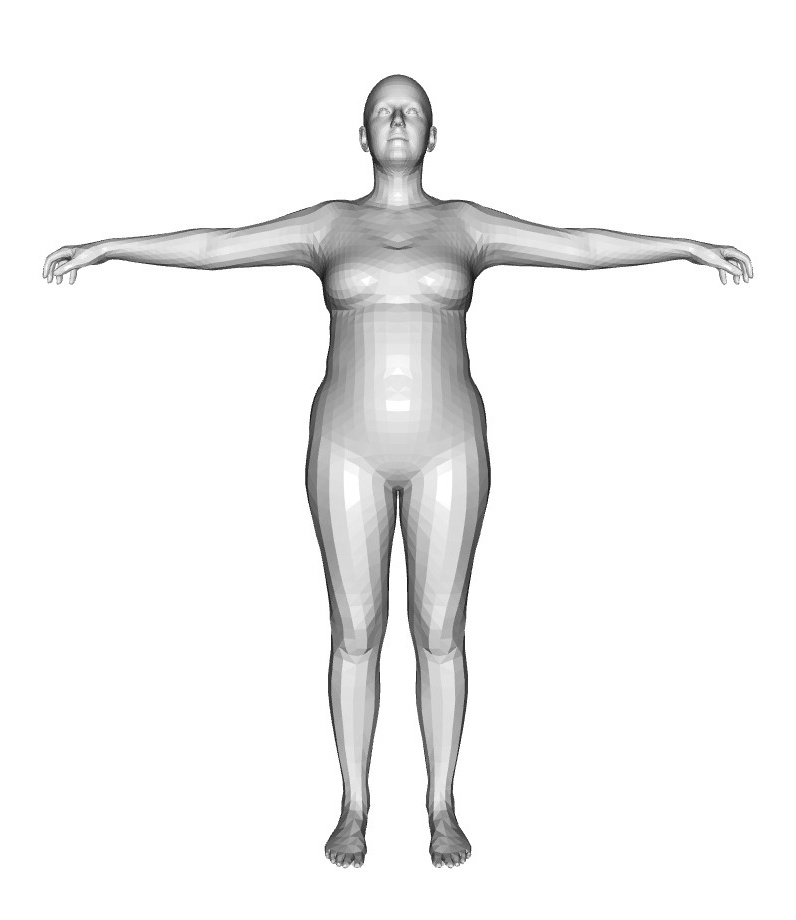} & 
        \includegraphics[width=\lw cm]{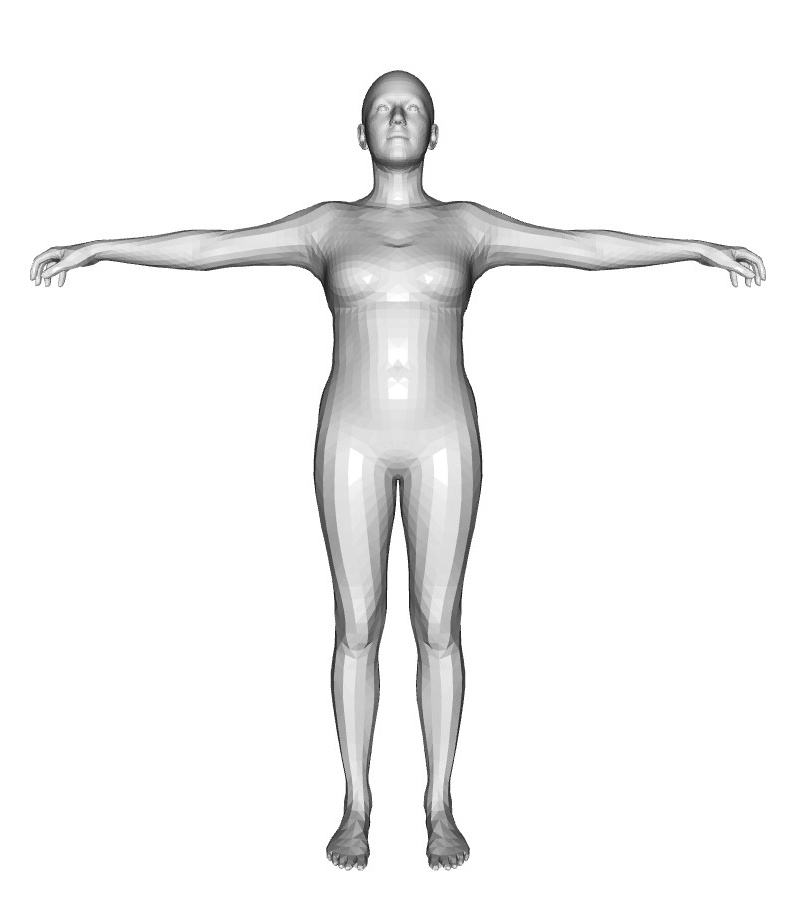} &&
        \includegraphics[width=\lw cm]{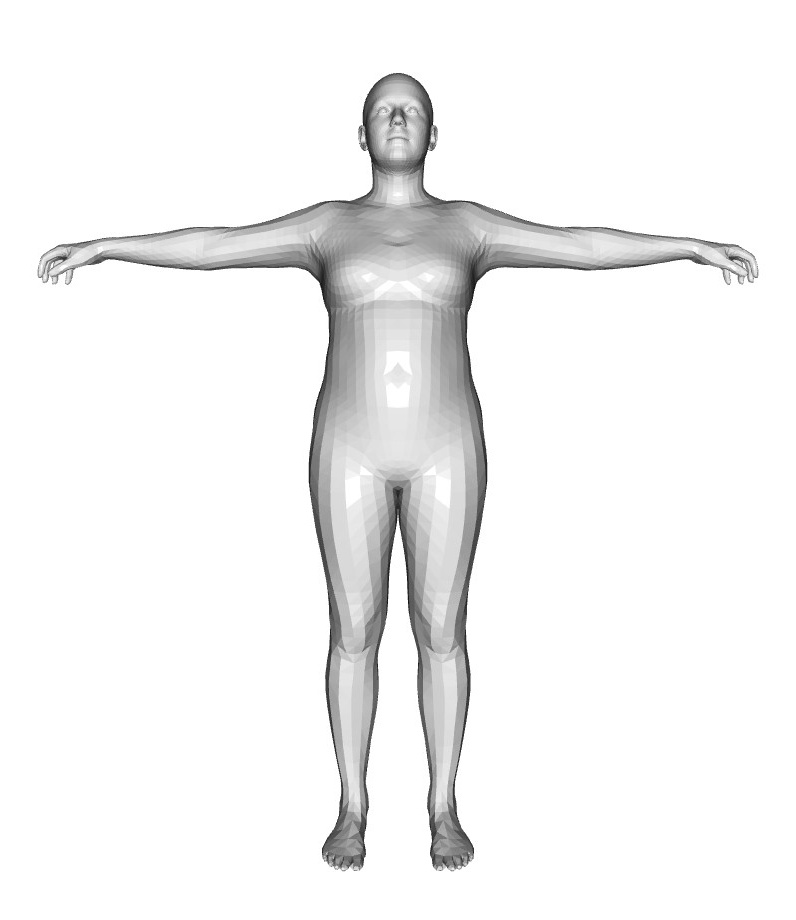} & 
        \includegraphics[width=\lw cm]{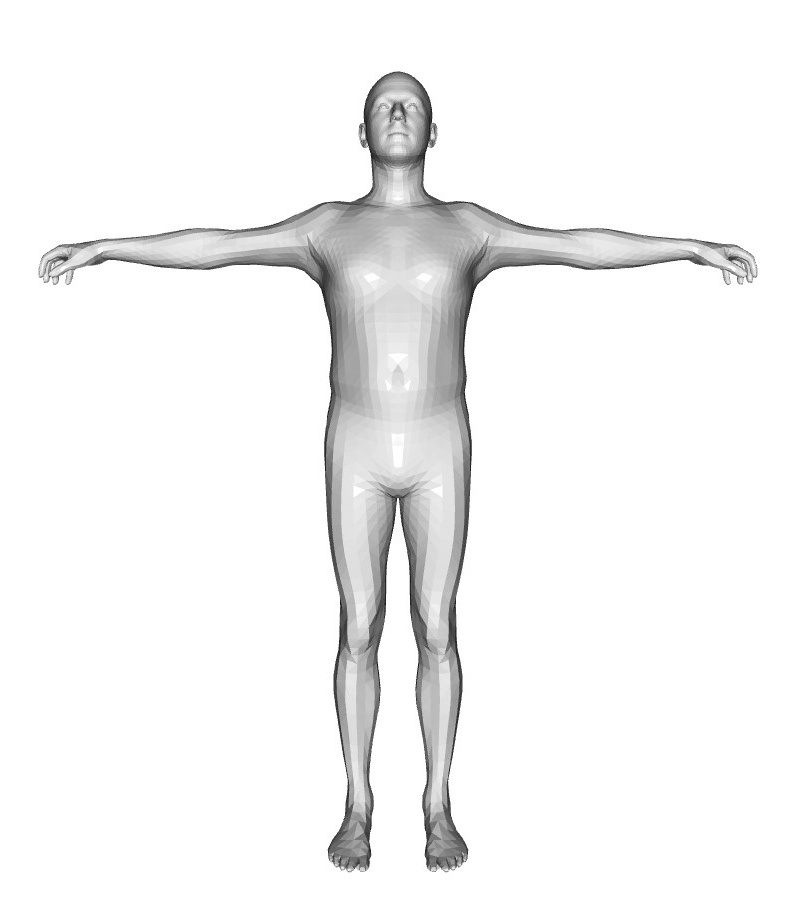}\\

        \includegraphics[width=\lw cm]{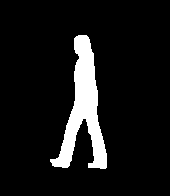} &&
        \includegraphics[width=\lw cm]{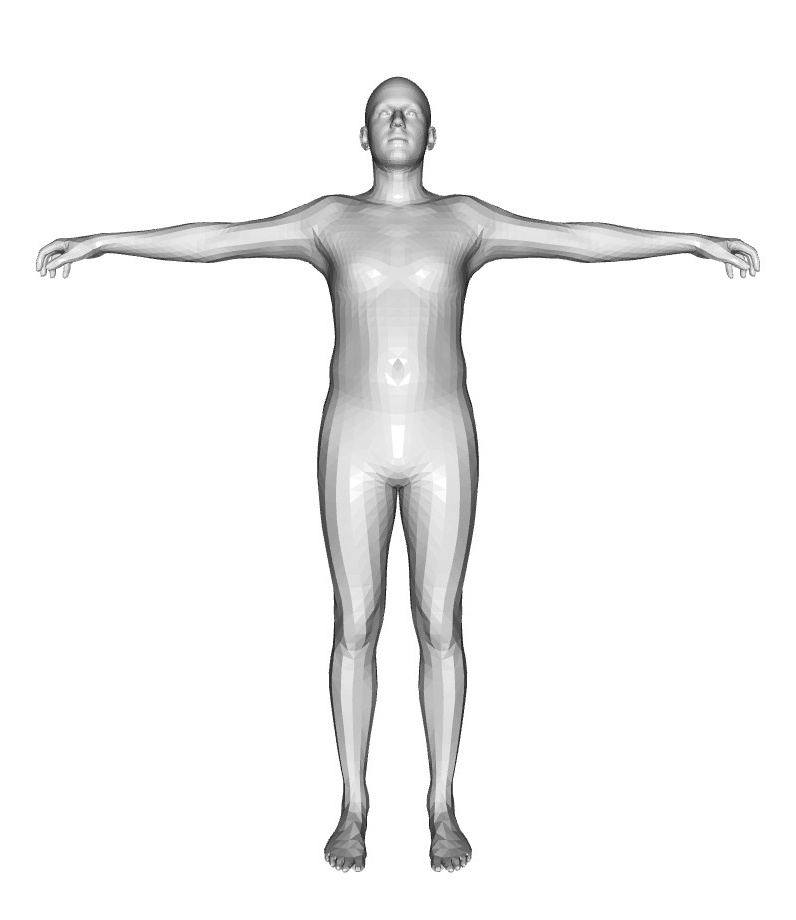} & 
        \includegraphics[width=\lw cm]{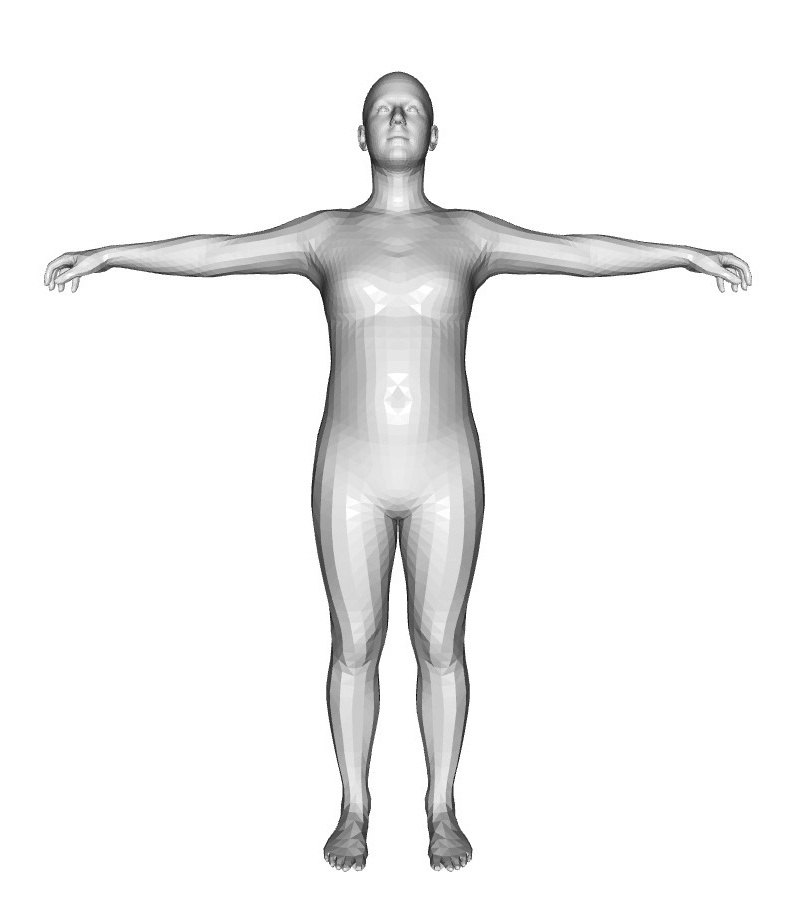} &&
        \includegraphics[width=\lw cm]{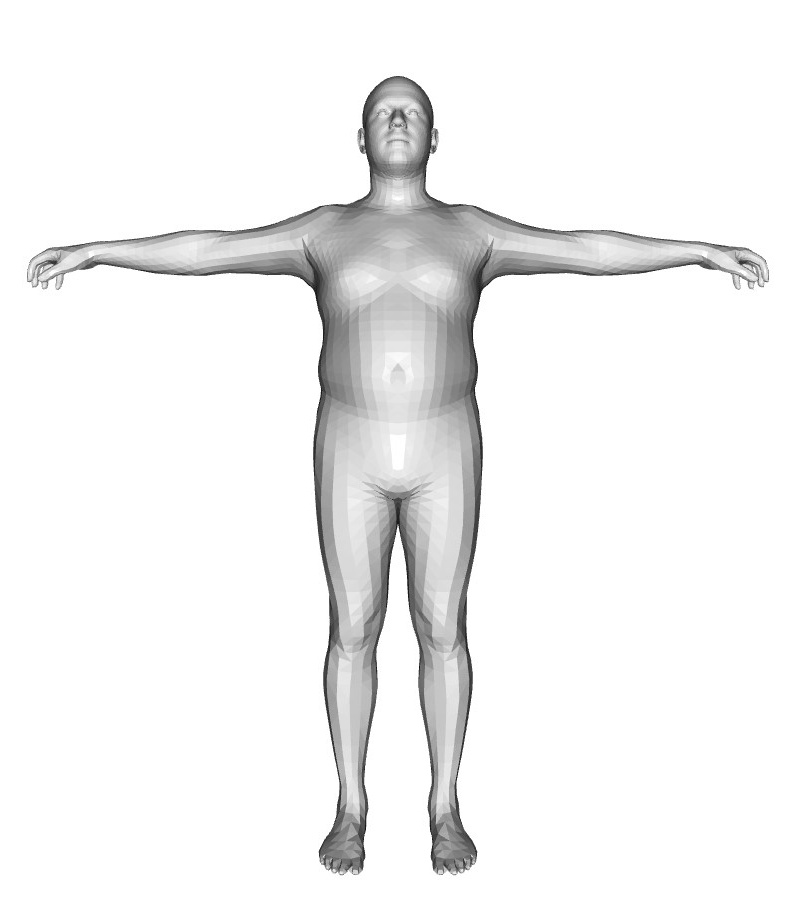} & 
        \includegraphics[width=\lw cm]{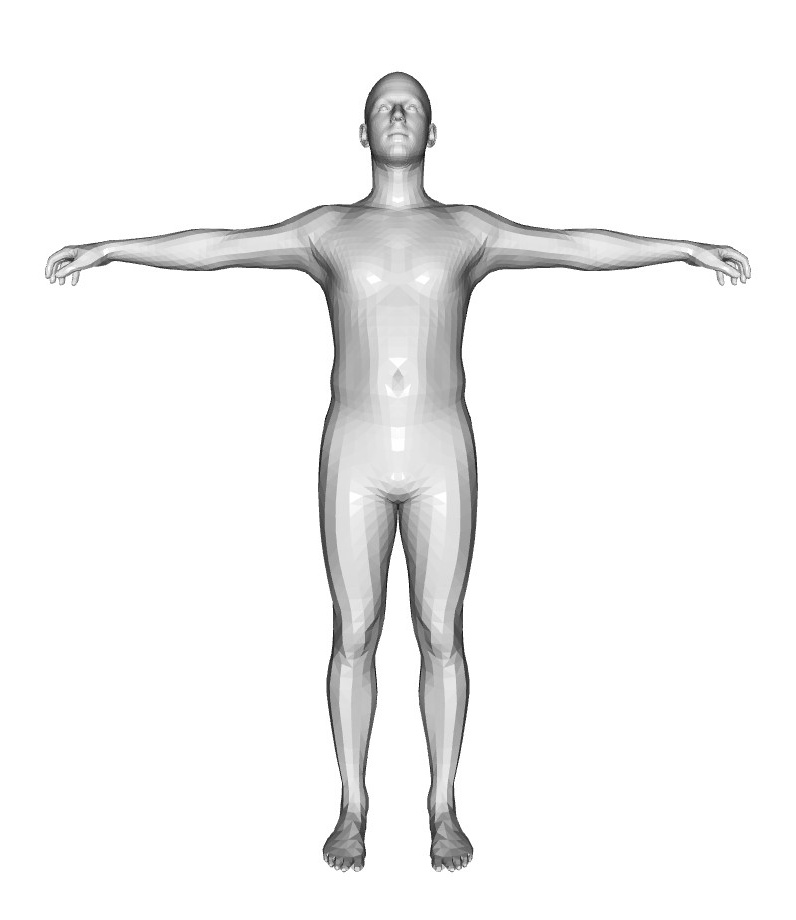} &&
        \includegraphics[width=\lw cm]{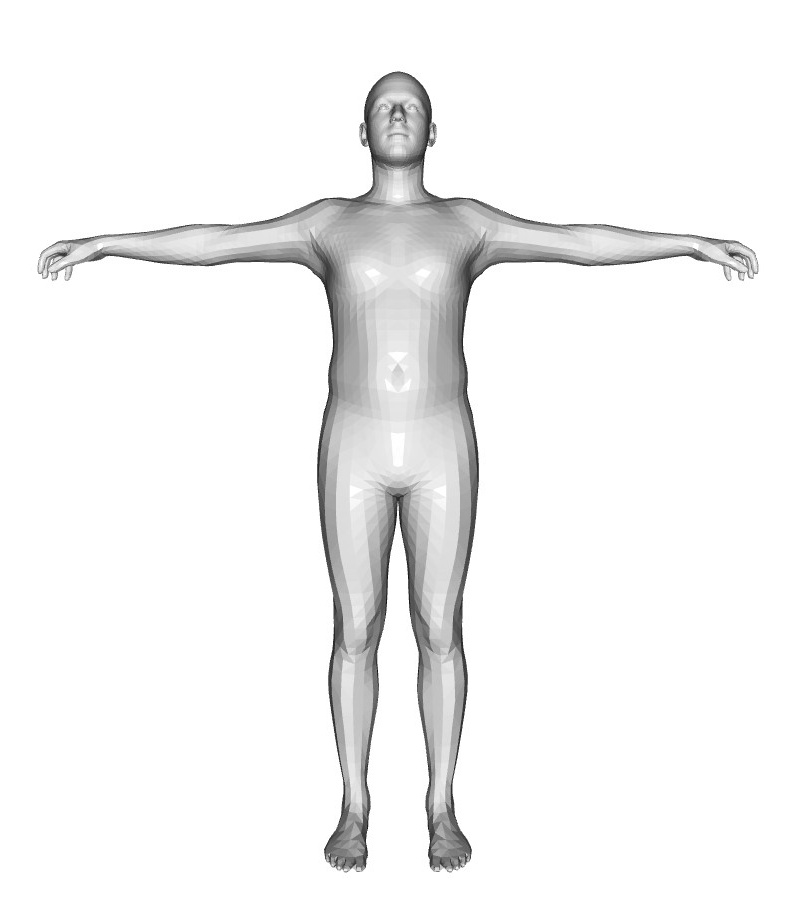} & 
        \includegraphics[width=\lw cm]{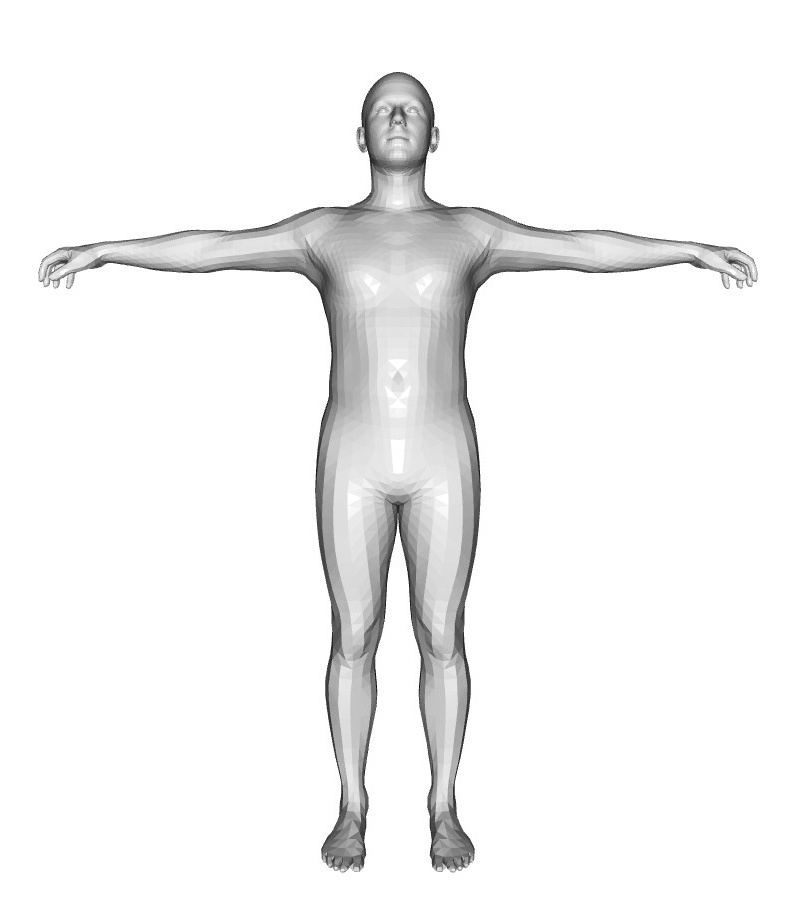}\\
        
        \bottomrule
\end{tabular}
}
\medskip
\caption{Visualizations for reconstructed human body shapes of two identities from selected RGB frames and silhouettes in the CASIA-B test set. For each example, the camera position from top to down is 0\degree, 36\degree, 72\degree\ and 108\degree\ respectively. We align the camera position to the front view for all variations and plot T-pose shapes for each person with the $\beta$ we inferred from the human body shape encoder. `RGB' and `silhouette' represent the reconstruction is from the branch with selected RGB images (SMPLify-X \cite{SMPL-X:2019}) or the gait feature extraction branch (Body Shape Feature Encoder). Silhouettes shown in the first column only indicate the IDs of the people and camera viewpoints, which are not the sequences used for body shape reconstruction.}
\label{fig:Visualizations}
\end{figure*}

We show the visualizations of inferred body shapes in Fig.~\ref{fig:Visualizations}, along with one of the silhouettes sampled at each camera viewpoint. 
We note that reconstructions from both methods, SMPLify-X \cite{SMPL-X:2019} and our body shape feature encoder, are pretty accurate for reconstructing human body shapes in the selected frames or silhouettes. For example, the first person is broader than the second, which can be reflected in most reconstructed meshes. In addition, both reconstructed shapes show good robustness again different appearance variations and different viewpoints, while shapes reconstructed from silhouette sequences by our body shape feature encoder are more consistent for the same person. Compared with a single frame of selected RGB images, a sequence input gives more information for reconstructing the human body shape and is more precise in describing the shape using information from neighbor frames.

\begin{figure}[t]
\centering
\def\lw{3}
\def\ls{0.5}
\resizebox{0.95\linewidth}{!}
{
\centering
\begin{tabular}{p{\lw cm}<{\centering}p{\ls cm}<{\centering}p{\lw cm}<{\centering}}
        \includegraphics[height=2.7 cm]{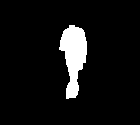}&&
        \includegraphics[height=2.7 cm]{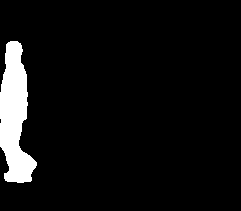}\\
        \small{(a) Incomplete cases} && \small{(b) Boundary cases}\\
\end{tabular}
}
\caption{Sampled silhouette visualization for error prediction.}
\label{fig:bad cases}
\end{figure}

\section{Limitation and Error Analysis}
 To distill and transfer knowledge from limited RGB images to the body shape feature encoder of the gait branch, we use SMPLify-X \cite{SMPL-X:2019} as our body prior extraction model for providing body shapes. The quality of the generated body prior from SMPLify-X is important. Although the distillation network is able to correct some mistakes generated from SMPLify-X as Figure~\ref{fig:Visualizations}, if there are too many mistakes from SMPLify-X, the distillation model will be unable to generate any useful body shapes for the training of body shape encoder in the gait branch.

During inference, our model has only one input, silhouette sequences. We note that the incomplete gait images, either from bad segmentation results or the person walking to the boundary of the image, as shown in Figure~\ref{fig:bad cases}, increase the probability of error prediction. When these incomplete silhouette images take a relatively large part of the video, the model is more likely to give wrong predictions since the silhouette is the only modality we have during inference.

\end{document}